\pdfoutput=1

\documentclass[11pt]{article}

\usepackage[final]{acl}

\usepackage{times}
\usepackage{latexsym}

\usepackage[T1]{fontenc}

\usepackage[utf8]{inputenc}

\usepackage{microtype}

\usepackage{inconsolata}

\usepackage{graphicx}

\usepackage{caption}
\usepackage{graphicx}

\title{Instructions for *ACL Proceedings}

\author{Xiaoyang Hu \\
  Brown University \\
  \texttt{xiaoyang\_hu@brown.edu} \\\And
  Richard L. Lewis \\
  University of Michigan \\
  \texttt{rickl@umich.edu} \\}

\usepackage{amsmath}
\usepackage{tikz}
\newcommand{\roundedbox}[1]{%
    \tikz\node[draw, rounded corners, inner sep=6pt]{#1};%
}
\usepackage{amsmath}
\usepackage{amssymb}
\usepackage{dsfont}
\usepackage{setspace}
\usepackage{enumitem}
\usepackage{microtype}
\usepackage{multirow}
\usepackage{multicol}
\usepackage{booktabs}
\usepackage{array}
\usepackage{listings}
\usepackage{xcolor}
\usepackage{upquote}
\usepackage{float}

\title{Do Language Models Understand the Cognitive Tasks Given to Them?\\Investigations with the \textit{N}-Back Paradigm}

\date{}

\begin{document}
\maketitle
\begin{abstract}
Cognitive tasks originally developed for humans are now increasingly used to study language models. While applying these tasks is often straightforward, interpreting their results can be challenging. In particular, when a model underperforms, it is often unclear whether this results from a limitation in the cognitive ability being tested or a failure to understand the task itself.
A recent study argues that \textsc{GPT 3.5}'s declining performance on 2-back and 3-back tasks reflects a working memory capacity limit similar to humans \citep{gong2024working}.
By analyzing a range of open-source language models of varying performance levels on these tasks, we show that the poor performance is due at least in part to a limitation in task comprehension and task set maintenance. We challenge the best-performing model with progressively harder versions of the task (up to 10-back) and experiment with alternative prompting strategies, before analyzing model attentions. Our larger aim is to contribute to the ongoing conversation around refining methodologies for the cognitive evaluation of language models.\footnote{Code available at \url{https://github.com/hxiaoyang/lm-nback}.}
\end{abstract}

\section{Introduction}
\begin{figure}[t]
  \centering
  \includegraphics[width=0.85\columnwidth]{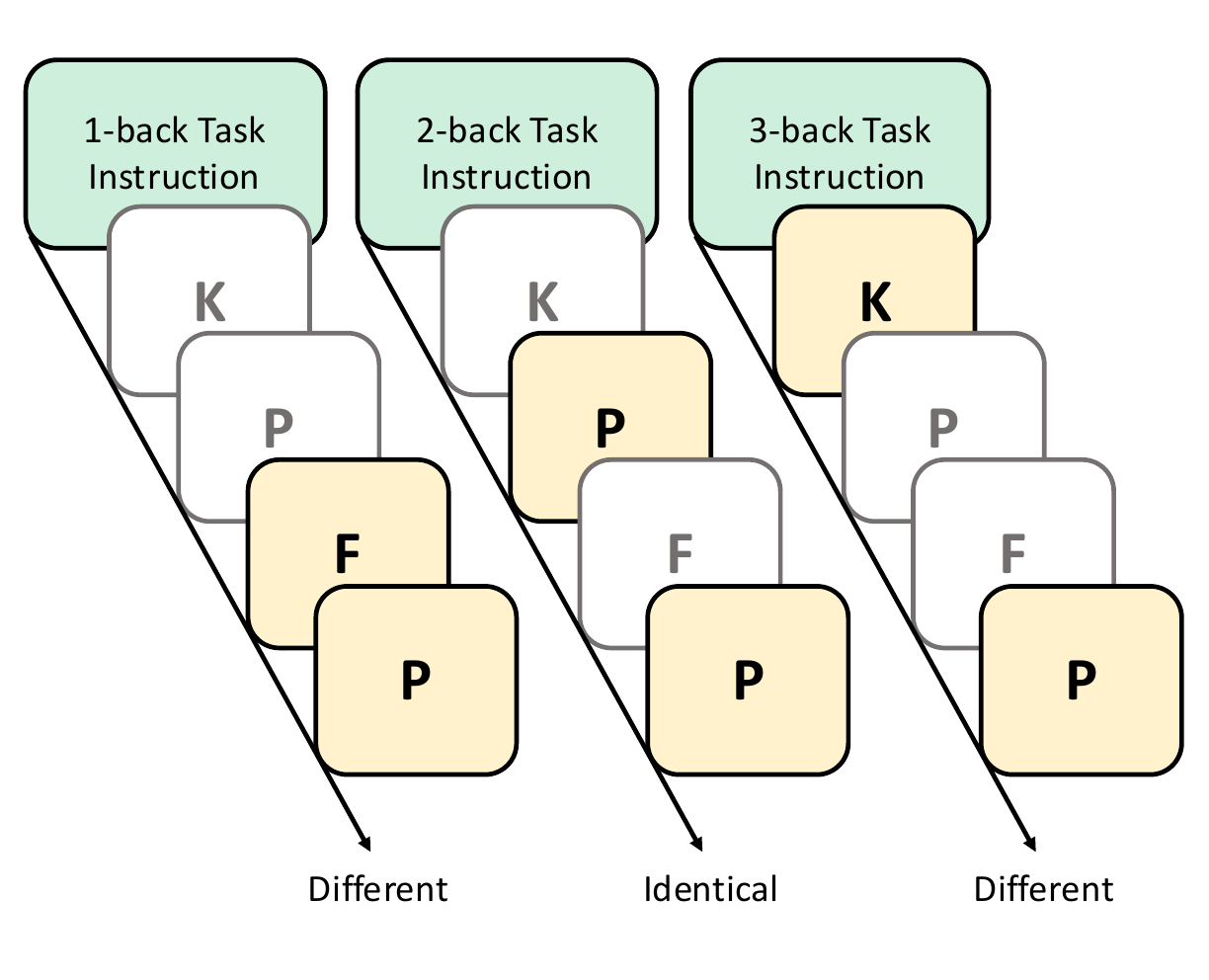}
  \caption{The $n$-back task is a common working memory task in which subjects are presented with a sequence of stimuli. At each step, they must decide whether the current item matches the one appearing $n$ step(s) earlier. This requires them to continuously update a list of $n$ most recent stimuli in the working memory.}
  \label{fig:nback_tasks}
\end{figure}

Psychologists rely on behavioral experiments to test hypotheses about cognitive constructs and processes. For these experiments to be valid, participants have to understand exactly what they are being asked to do. To that end, human study protocols often include detailed task instructions, demonstrations, and practice runs. When adapting these experiments for language models, ensuring task comprehension can be more challenging, given that these models are often more hesitant than humans to express uncertainty \citep{zhou2024relying}.

A recent study applies the $n$-back task (Figure~\ref{fig:nback_tasks}) to GPT 3.5 and concludes from the model's poor 2-back and 3-back performance that it has a working memory capacity limit (WMCL) of approximately 3, apparently similar to humans \citep{gong2024working}. This interpretation raises two concerns. First, while WMCL is well established in human cognition, we cannot assume these same constraints exist or can be meaningfully measured in language models. Second, these results may reflect the model's failure to understand the task requirements rather than any inherent memory limitation.

In this paper, we show that low-performing language models, even when provided with detailed $n$-back task instructions and demonstrations, commit errors that are consistent with a different $m$-back task ($m\neq n$). Notice that, if a human subject committed such systematic errors, we would conclude that they had misunderstood the task.
In comparison, intermediate models, including GPT 3.5, tend to start with the correct task but drift toward a different one as errors accumulate, resulting in poor average 2-back and 3-back performance, consistent with \citealt{gong2024working}.
High-performing models, on the other hand, consistently execute the correct task, even for larger $n$'s, achieving task accuracies of 90.08\%, 90.08\%, and 84.75\% for $n=8,9,10$.

The remainder of this paper is organized as follows.
Section~\ref{sec:background} covers relevant background and related work.
Section~\ref{sec:methods} introduces the dataset, models, prompting approach, and evaluation metrics.
Section~\ref{sec:task-performance} benchmarks each model on 1-back, 2-back, and 3-back tasks, identifying three distinct performance tiers.
Section~\ref{sec:task-comprehension} investigates whether these performance disparities are explained by differences in task comprehension.
Section~\ref{sec:task-set-maintenance} examines the models' ability to consistently apply the correct task set throughout each trial (task set maintenance).
In Section~\ref{sec:llama-70b-10bck}, we challenge the best model to perform 1-back through 10-back tasks and notice a signature of task comprehension.
In Sections~\ref{sec:curriculum-learning} and~\ref{sec:interactive-demo}, we discuss additional experiments with alternative prompting strategies for comparison.
In Section~\ref{sec:attn-analysis}, we identify an attention pattern whose prevalence predicts 2-back task performance.

\section{Background and Related Work}
\label{sec:background}
There has been a growing body of work that evaluates pre-trained language models using cognitive tasks originally developed for humans. These efforts often aim to identify whether the models exhibit cognitive constructs or capabilities that are present in humans. Subjects of study include theory of mind \citep{strachan2024testing, gandhi2024understanding}, analogical reasoning \citep{hu2023context, webb2023emergent}, cognitive biases \citep{binz2023using, lampinen2024language}, and WMCL \citep{gong2024working}, among many others. Such evaluations are susceptible to both overclaiming and underclaiming. On the one hand, false positives can result from training data contamination \citep{sainz2023nlp}, potentially compromising the validity of vignette-based assessments where models may produce memorized responses. On the other hand, underestimation of model capabilities can happen when we erroneously assume task comprehension, especially for smaller models \citep{hu2024auxiliary}. Prior studies have also investigated how well language models adhere to prompt instructions, especially compared to humans \citep{webson2022prompt, webson2023language}. In light of other methodological challenges in the cognitive evaluation of language models, such as prompt sensitivity and cultural biases, \citealt{ivanova2023running} outlines recommendations for best practices.

Virtually any task, from routine text comprehension to complex problem solving, involves the creation of intermediate or partial results. Successful task completion requires that these results be maintained in a way that facilitates later access. In humans, this mechanism is known as \textit{working memory}, one of the most studied constructs in psychology for over half a century \citep{miyake1999models}. This concept can be extended to transformer-based language models designed to process interdependent, serial information. In fact, the transformer architecture, particularly its attention mechanism where key-query matching drives retrieval \citep{vaswani2017attention}, bears striking resemblance to cue-based parsing and retrieval models proposed in psycholinguistics \citep{lewis2006computational}, making it a promising candidate for modeling human sentence processing. One of the most salient and mysterious aspects of human working memory is its severely constrained capacity \citep{miller1956magical, cowan2012working}. One prominent task used to measure working memory capacity is the $n$-back task \citep{kirchner1958age}.

To the best of our knowledge, \citealt{gong2024working} are the first to apply the $n$-back task to a language model, specifically the \textsc{GPT 3.5 Turbo} variant of ChatGPT. They experiment with different prompting strategies, including those incorporating feedback and reasoning. As $n$ increases from 1 to 3, they observe a sharp decline in model performance and conclude that the model has a WMCL of approximately 3.
{\color{black}{\citealt{zhang2024working} also examine working memory in language models using a task described as $n$-back, but with all stimuli presented simultaneously.
This departs from the standard paradigm and imposes different working memory demands.
Moreover, while they acknowledge that the poor performance in smaller models may stem from limited understanding of the ``intent of the input'', they do not control for task comprehension as a confounding variable.

\textit{Multi-hop question answering} is another working memory task paradigm, in which two or more reasoning steps must be performed sequentially to resolve complex queries.
This task is interestingly different from $n$-back in that it places \textit{implicit} demands on working memory in a single forward pass.
For instance, answering ``\textit{The spouse of the performer of Imagine was...}'' requires first identifying John Lennon as the performer of the song and then determining that Yoko Ono was his spouse.
\citealt{biran2024hopping} find that, after early model layers resolve the initial step, later layers often prove deficient in completing the second.

}}

\begin{figure*}[t]
  \includegraphics[width=1\linewidth]{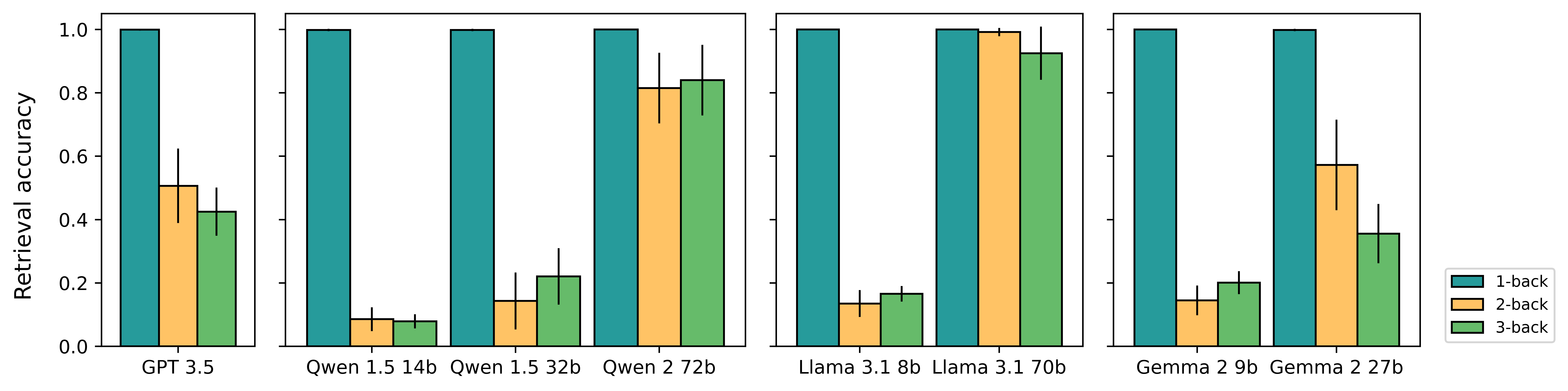}
  \caption {Average retrieval accuracies on 1-, 2-, and 3-back tasks, grouped by model family.}
  \label{fig:perf-v1}
\end{figure*}

\section{Methods}
\label{sec:methods}

\subsection{Data and Prompts}

We use the dataset from \citealt{gong2024working} (MIT License). For each $n$-back task, there are 50 trials in total. Each trial consists of a sequence of 24 letters. In exactly 8 random positions within each sequence, the letters are the same as those appearing $n$ step(s) earlier. After each letter prompt, the models are instructed to answer ``\{\textit{current letter}\}\textit{ and }\{\textit{letter $n$ back}\}\textit{ are }\{\textit{different / identical}\}'' This is designed to facilitate chain-of-thought reasoning \citep{wei2022chain} and to make explicit the specific letter retrieved by the model for comparison with the current one.
\begin{equation*}
\roundedbox{
\footnotesize
$
\begin{aligned}
&\left.\begin{aligned}
\texttt{SYS: } & \texttt{[TASK INSTRUCTIONS]} \\[-0.7ex]
\texttt{USR: } & \texttt{k} \\[-0.7ex]
\texttt{LLM: } & \texttt{k and none are different.} \\[-0.7ex]
\texttt{USR: } & \texttt{k} \\[-0.7ex]
\texttt{LLM: } & \texttt{k and k are identical.} \\[-0.7ex]
\texttt{USR: } & \texttt{a} \\[-0.7ex]
\texttt{LLM: } & \texttt{a and k are different.} \\[-0.7ex]
& \vdots
\end{aligned}\right\} \texttt{DEMO} \\[1ex]
&\left.\begin{aligned}
\texttt{SYS: } & \texttt{[TASK INSTRUCTIONS]} \\[-0.7ex]
\texttt{USR: } & \texttt{e} \\[-0.7ex]
\texttt{LLM: } & \underline{\texttt{e and none are different.}} \\[-0.7ex]
\texttt{USR: } & \texttt{f} \\[-0.7ex]
\texttt{LLM: } & \underline{\texttt{f and e are different.}} \\[-0.7ex]
\texttt{USR: } & \texttt{f} \\[-0.7ex]
\texttt{LLM: } & \underline{\texttt{f and f are identical.}} \\
& \vdots
\end{aligned}\right\} \texttt{TEST}
\end{aligned}
$}
\end{equation*}

To teach the models the correct answer format and maximize their chances of correctly inferring the tasks, each trial begins with a demonstration, which includes a sequence of 24 letters and the correct responses. The ``without demo'' trials in Section~\ref{sec:task-comprehension} are the only exception. Following the demonstration, a new sequence of 24 letters is presented, one at a time, and the models are prompted to respond after each letter. An example 1-back trial is shown above; actual model responses are underlined.

\subsection{Models}

We use \textsc{GPT 3.5 Turbo} and open-source instruction-tuned models from the \textsc{Qwen} \citep{bai2023qwen}, \textsc{Llama} \citep{dubey2024llama}, and \textsc{Gemma} \citep{team2024gemma} families. Each model is prompted recursively to complete the trials. For the open-source models, we analyze the token log probabilities and attention patterns in addition to the generated responses.

\subsection{Metrics}

The $n$-back task requires continuously matching the current letter and the letter from $n$ steps back to determine the correct label. Compared to binary labels, the retrieved letters offer better insight into the models' understanding of the task. And since the correct label is almost always assigned given the correct retrieval, our analyses focus on the retrieval accuracies and the log probabilities of the retrieved letters. One possibility for low performance is that, despite being prompted to do the $n$-back task, a model might be following $m$-back instructions instead. To investigate this, we adopt \textit{counterfactual} measures by providing $n$-back instructions and evaluating the accuracies and log probabilities of retrievals consistent with the $m$-back task. We also apply variants of these measures, which we detail in later sections.

\begin{table}
    \centering
    \footnotesize
    \setlength\tabcolsep{5pt}
    \begin{tabular}{c|c|c|c|c}
        \toprule
        \textbf{Tier} & \textbf{Model} & \textbf{1bk} & \textbf{2bk} & \textbf{3bk} \\
        \midrule
        \multirow{4}{*}{T3}
        & \textsc{Qwen 1.5 14b Chat} & 1.00 & 0.09 & 0.08 \\
        & \textsc{Llama 3.1 8b Instr.} & 1.00 & 0.14 & 0.17 \\
        & \textsc{Gemma 2 9b Instr.} & 1.00 & 0.15 & 0.20 \\
        & \textsc{Qwen 1.5 32b Chat} & 1.00 & 0.14 & 0.22 \\
        \midrule
        \multirow{2}{*}{T2}
        & \textsc{Gemma 2 27b Instr.} & 1.00 & 0.57 & 0.36 \\
        & \textsc{GPT 3.5 Turbo} & 1.00 & 0.51 & 0.43 \\
        \midrule
        \multirow{2}{*}{T1} 
        & \textsc{Qwen 2 72b Instr.} & 1.00 & 0.81 & 0.84 \\
        & \textsc{Llama 3.1 70b Instr.} & 1.00 & 0.99 & 0.93 \\
        \bottomrule
    \end{tabular}
    \normalsize
    \caption{Average retrieval accuracies on 1-, 2-, and 3-back tasks, organized by performance tier.}
    \label{tab:models-by-tier}
\end{table}

\section{Experimental Results}\label{sec:results}

\subsection{Task Performance}
\label{sec:task-performance}

We begin by comparing retrieval accuracies across models for all three tasks (Figure~\ref{fig:perf-v1}) and categorizing them into three performance tiers (Table~\ref{tab:models-by-tier}).

T3: $\leq 20\%$ on 2- and 3-back.

T2: $\sim 50\%$ on 2-back; $\sim 40\%$ on 3-back.

T1: $> 80\%$ on 2- and 3-back.

For subsequent analyses, we select the best-performing model, \textsc{Llama 3.1 70b Instruct} (T1), the worst-performing model, \textsc{Qwen 1.5 14b Chat} (T3), and \textsc{Gemma 2 27b Instruct} (T2) to represent each performance tier.

\begin{figure}[H]
  \centering
  \includegraphics[width=0.85\columnwidth]{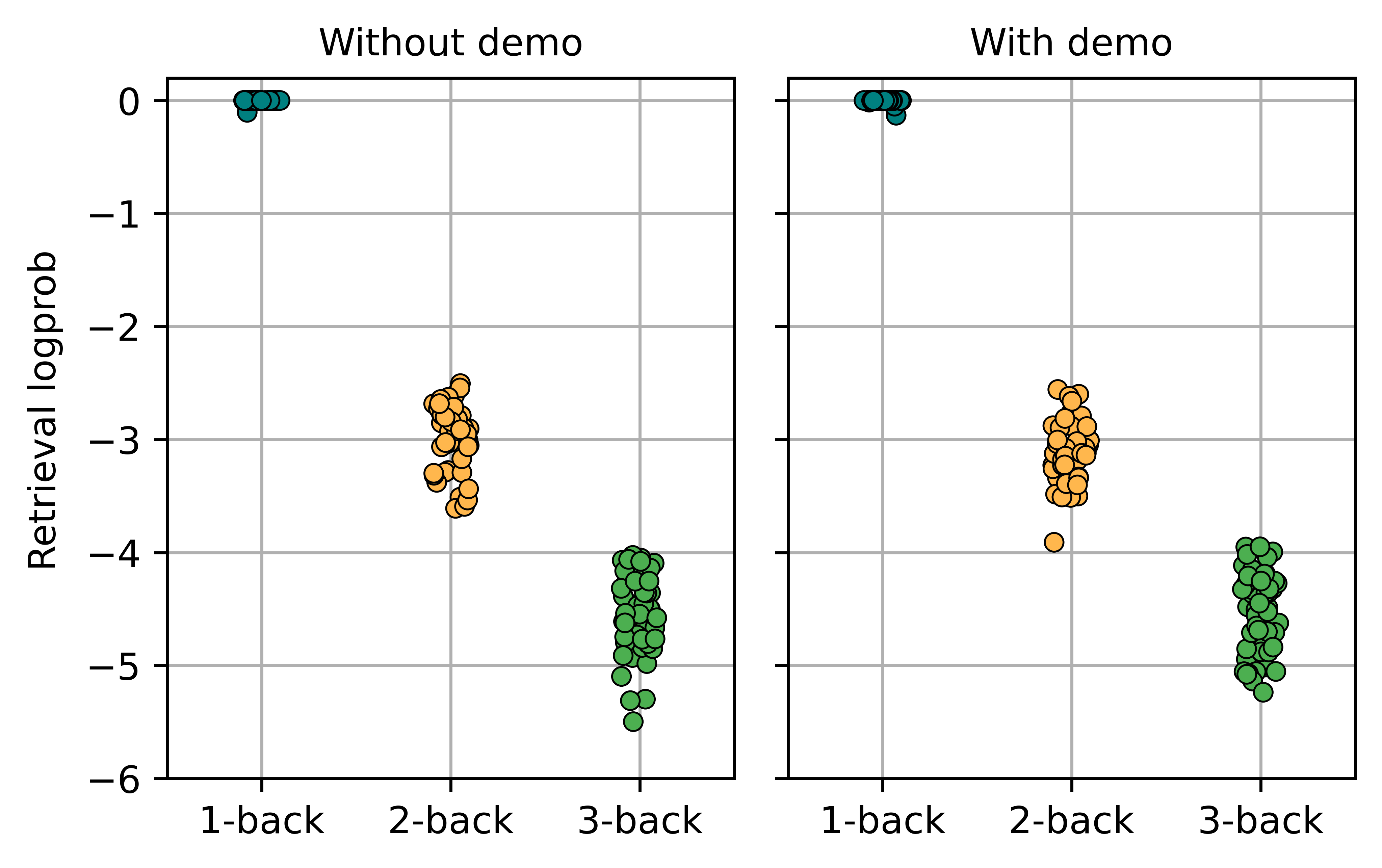}
  \includegraphics[width=0.85\columnwidth]{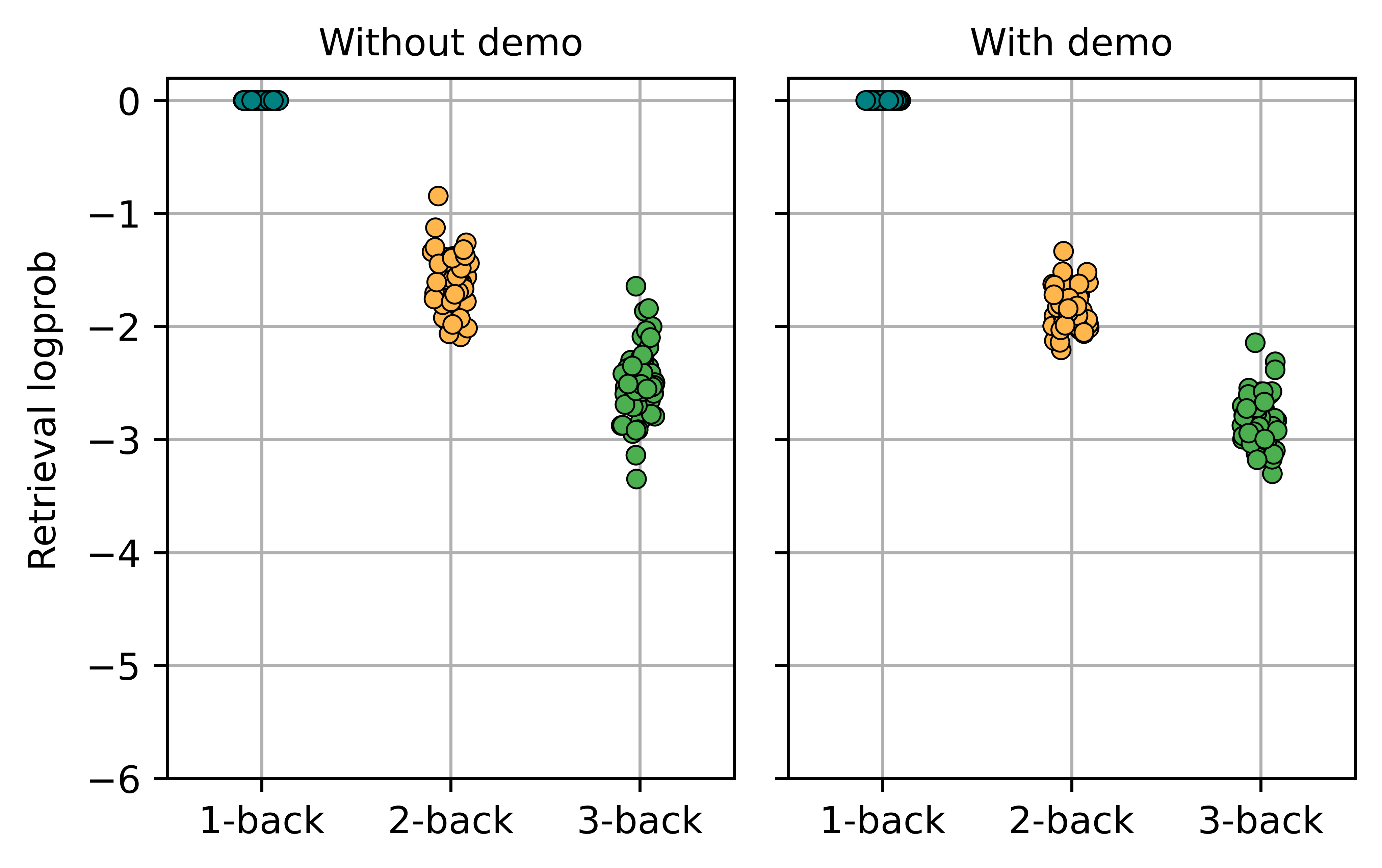}
  \includegraphics[width=0.85\columnwidth]{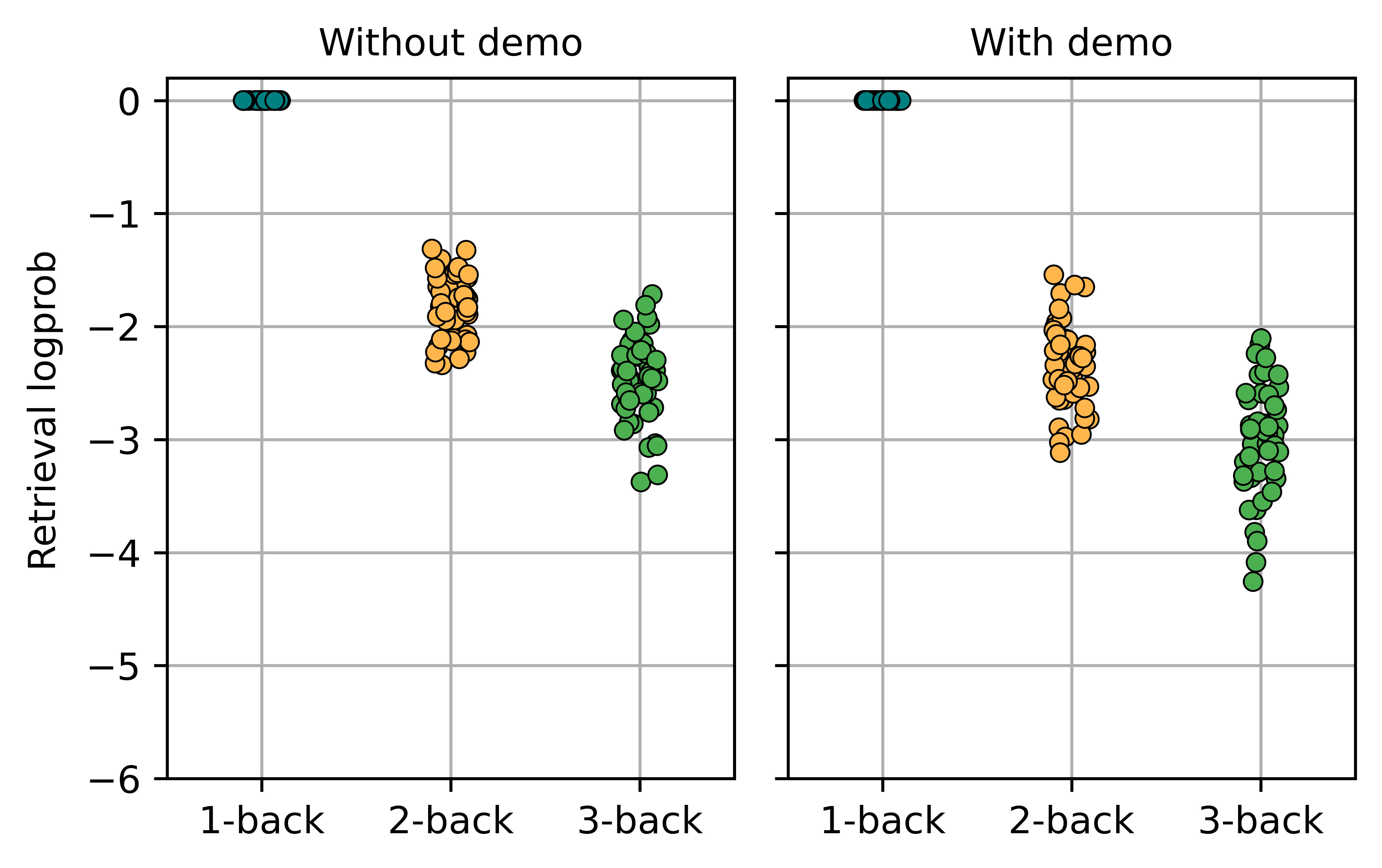}
  \caption{Retrieval log probabilities for 1-back task continuations, with and without demonstrations. From top to bottom are results for \textsc{Qwen 1.5 14b Chat} (T3), \textsc{Gemma 2 27b Instruct} (T2), and \textsc{Llama 3.1 70b Instruct (T1)}. Each point corresponds to the average retrieval log probability of one trial.}
  \label{fig:1back-strip}
\end{figure}

\subsection{Task Comprehension}
\label{sec:task-comprehension}

To better understand the source of these performance disparities, we ask:
Can models infer the task from the instructions and demonstrations—and if so, which serves as a more effective cue?
To address these questions, we 1) provide $n$-back instructions with and without demonstrations, 2) present three continuations, each consistent with a different $m$-back task, and 3) measure the average retrieval log probabilities for each trial.

Let $\mathsf{P}^{-}_{n,m}$ be the average $m$-back retrieval log probability given $n$-back instructions only. Let $\mathsf{P}_{n,m}$ be the average $m$-back retrieval log probability given $n$-back instructions and demonstrations.

\paragraph{1-back.}
Under 1-back instructions, $\mathsf{P}_{1,1}>\mathsf{P}_{1,2}>\mathsf{P}_{1,3}$ across all models. The same is true when no task demonstrations are provided, with no significant difference between $\mathsf{P}_{1,m}$ and $\mathsf{P}^-_{1,m}$ for $m=1,2,3,$ as shown in Figure~\ref{fig:1back-strip}. Overall, this is unsurprising, given the near-perfect performances of all models on 1-back trials.

\begin{figure}[t]
  \centering
  \includegraphics[width=0.85\columnwidth]{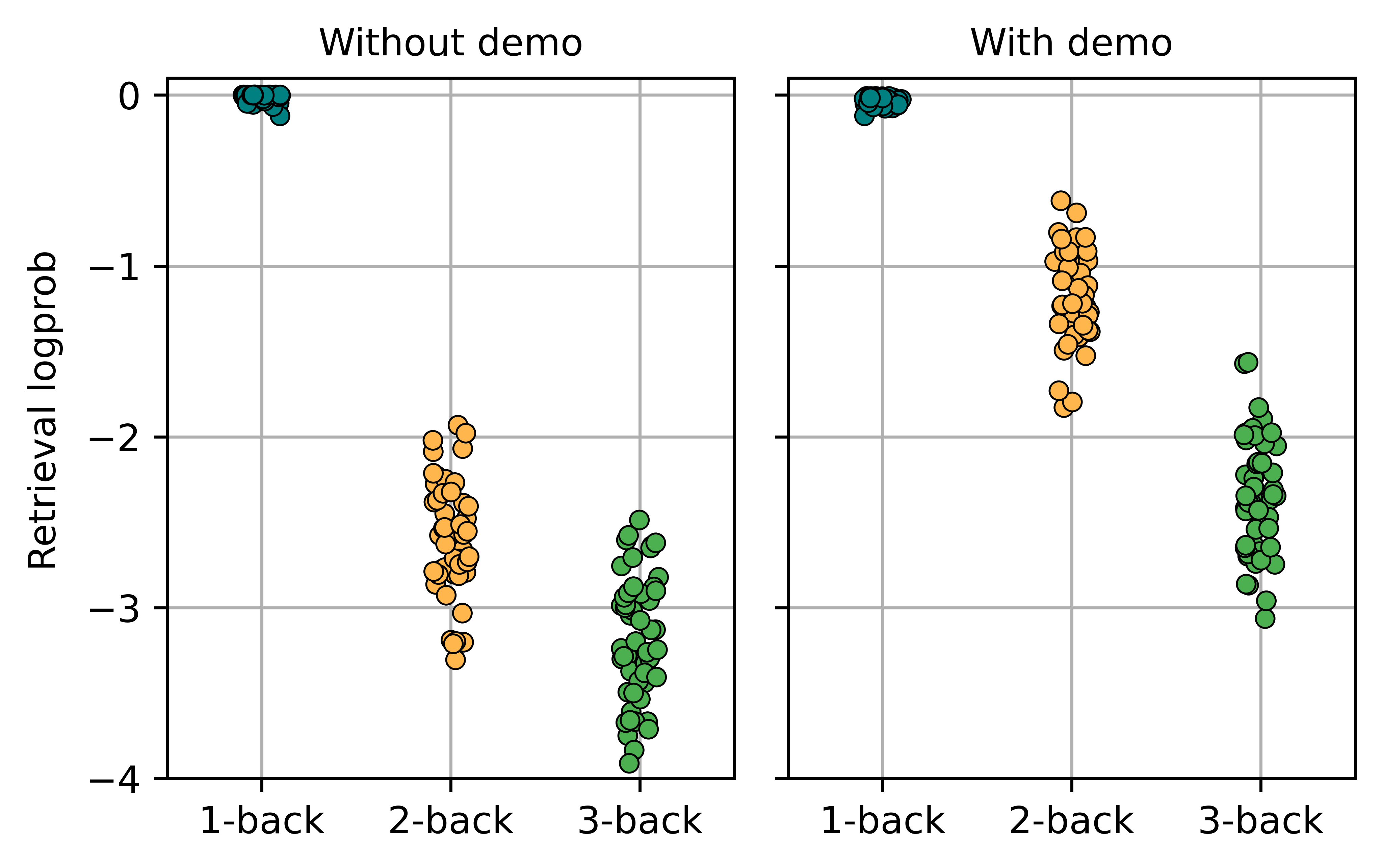}
  \includegraphics[width=0.85\columnwidth]{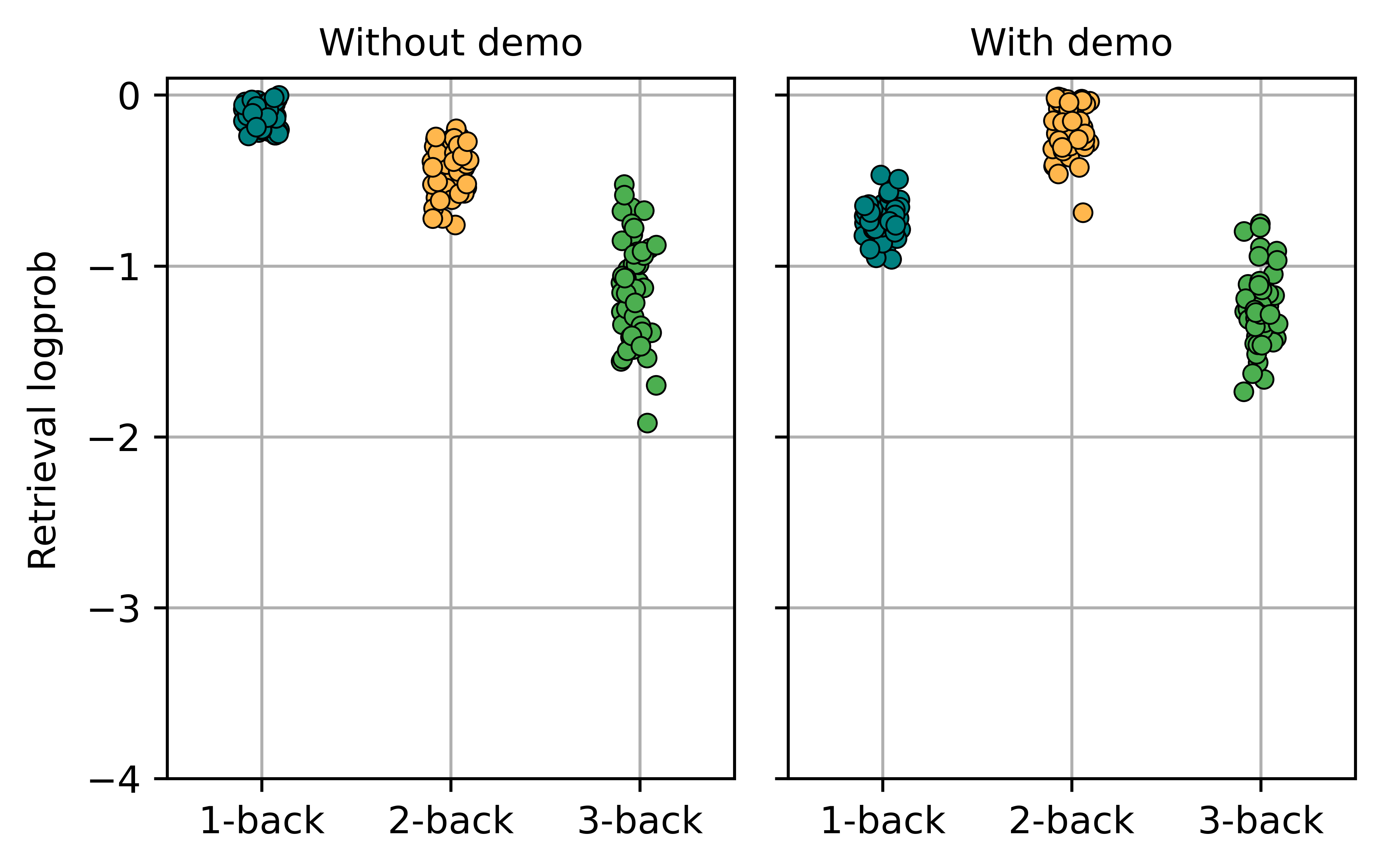}
  \includegraphics[width=0.85\columnwidth]{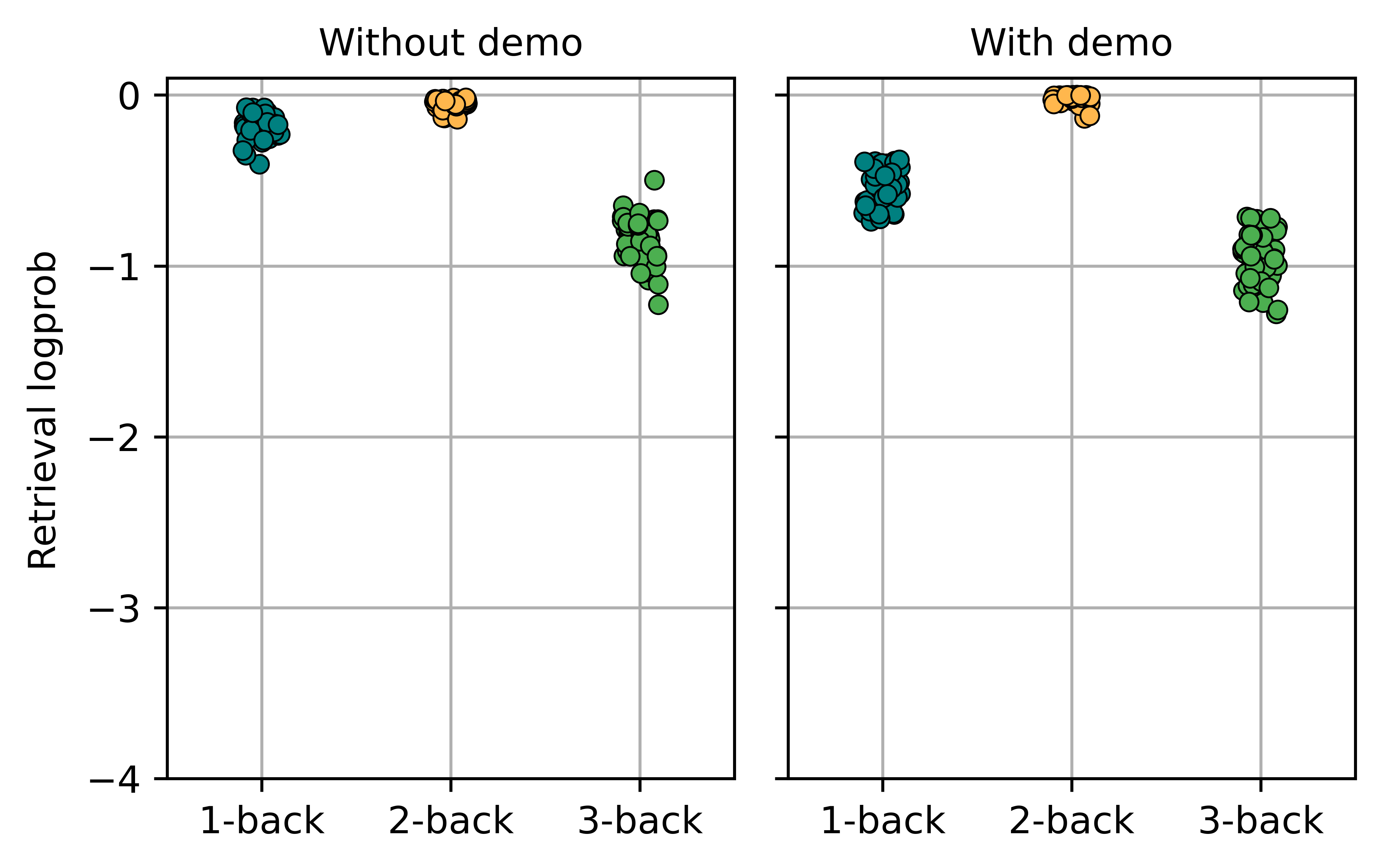}
  \caption{Retrieval log probabilities for 2-back task continuations, with and without demonstrations. From top to bottom are results for \textsc{Qwen 1.5 14b Chat} (T3), \textsc{Gemma 2 27b Instruct} (T2), and \textsc{Llama 3.1 70b Instruct} (T1). Each point corresponds to the average retrieval log probability of one trial.}
  \label{fig:2back-strip}
\end{figure}

\begin{figure}[t]
  \centering
  \includegraphics[width=0.85\columnwidth]{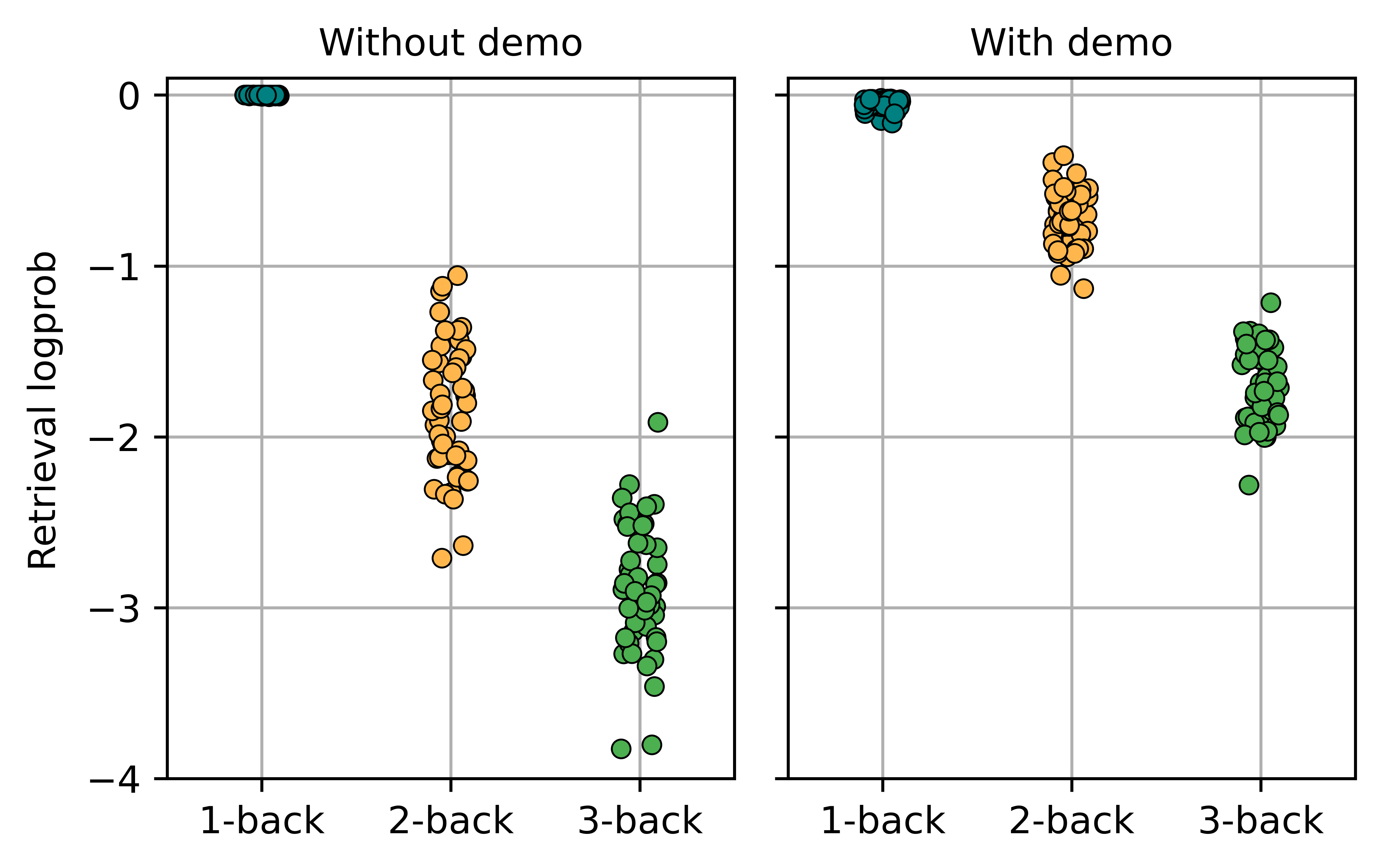}
  \includegraphics[width=0.85\columnwidth]{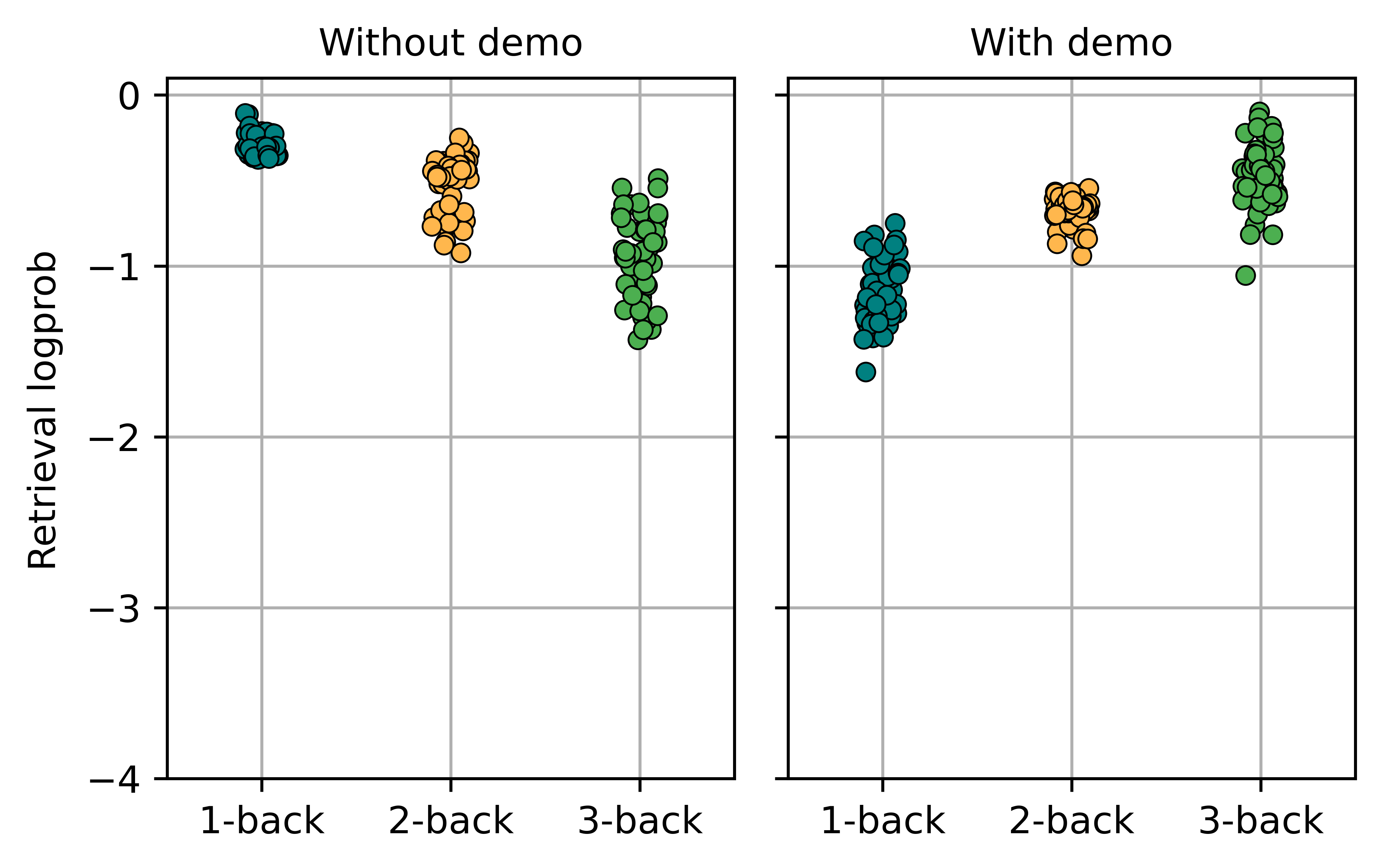}
  \includegraphics[width=0.85\columnwidth]{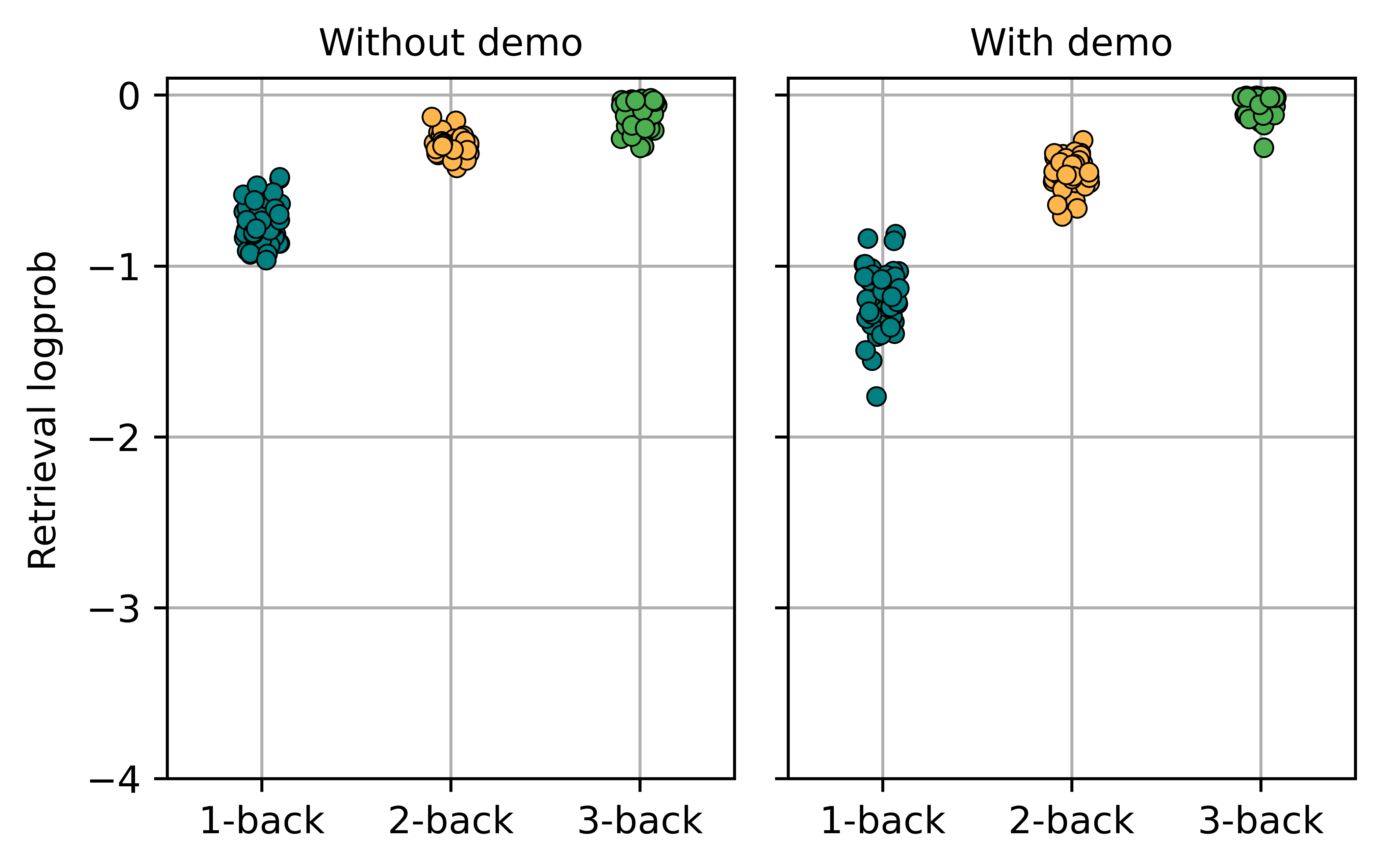}
  \caption{Retrieval log probabilities for 3-back task continuations, with and without demonstrations. From top to bottom are results for \textsc{Qwen 1.5 14b Chat} (T3), \textsc{Gemma 2 27b Instruct} (T2), and \textsc{Llama 3.1 70b Instruct} (T1). Each point corresponds to the average retrieval log probability of one trial.}
  \label{fig:3back-strip}
\end{figure}

\paragraph{2-back.}
We analyze the representative model from each tier (Figure~\ref{fig:2back-strip}).

T3: Under 2-back instructions, including with demonstrations, 1-back continuations are assigned to be the most plausible, with both $\mathsf{P}^-_{2,1}>\mathsf{P}^-_{2,2}>\mathsf{P}^-_{2,3}$ and $\mathsf{P}_{2,1}>\mathsf{P}_{2,2}>\mathsf{P}_{2,3}$. The task demonstrations do bring $\mathsf{P}_{2,2}$ and $\mathsf{P}_{2,3}$ closer to $\mathsf{P}_{2,1}$, although this is not enough to offset the strong 1-back priors.

T2: Under 2-back instructions only, the ordering of $\mathsf{P}^-_{2,m}$ remains the same, albeit with $\mathsf{P}^-_{2,2}$ and $\mathsf{P}^-_{2,3}$ noticeably closer to $\mathsf{P}^-_{2,1}$ than for T3. However, with additional task demonstrations, 2-back continuations are assigned to be the most likely, with $\mathsf{P}_{2,2}>\mathsf{P}_{2,1}>\mathsf{P}_{2,3}$.

T1: Somewhat surprisingly, we notice that $\mathsf{P}^-_{2,2}>\mathsf{P}^-_{2,1}>\mathsf{P}^-_{2,3}$, showing that the model is able to infer the task from the instructions alone. However, the demonstrations do help further consolidate the mapping.

\paragraph{3-back.}
As shown in Figure~\ref{fig:3back-strip}, the 3-back patterns are largely analogous to the 2-back case.

\paragraph{Summary.}
{\color{black}{
Through analyzing models from different performance tiers, we identify three distinct levels of task comprehension capabilities. The T3 model fails to map 2-back and 3-back instructions to the correct responses, given either the instructions or demonstrations, suggesting it completely misunderstands the task; the T2 model fails to map 2-back and 3-back instructions to the correct responses, given the instructions, but can do so if demonstrations are also provided; the T1 model can map 2-back and 3-back instructions to the correct responses based on the instructions alone, suggesting a robust understanding of the tasks. The T3 model's failure to understand the task is corroborated by analyses in Section~\ref{sec:interactive-demo}. Even when provided with short demo sequences and immediate corrective feedback, it still fails to get 2 consecutive correct responses. This suggests that its poor performance stems from an inability to understand the task, rather than any memory limitation.
}}

\subsection{Task Set Maintenance}
\label{sec:task-set-maintenance}

Each $n$-back trial consists of a sequence of 24 letters. Successful task completion requires consistent adherence to the task instructions as more stimuli are presented. Here, we investigate whether language models show a progressive decline in their ability to produce $n$-back consistent responses over time. Previously, performance metrics were averaged across time steps for each trial. Now, we average across trials for each time step. At each time step $i$ in the $n$-back task, we measure the average accuracy of $m$-back consistent retrievals for each $m\leq n$, given the model's own responses up to time step $i-1$. Denote this as $\mathsf{A}_{n,\cdot}(m,i)$.

\begin{figure}[t]
    \centering
    \includegraphics[width=0.4\textwidth]{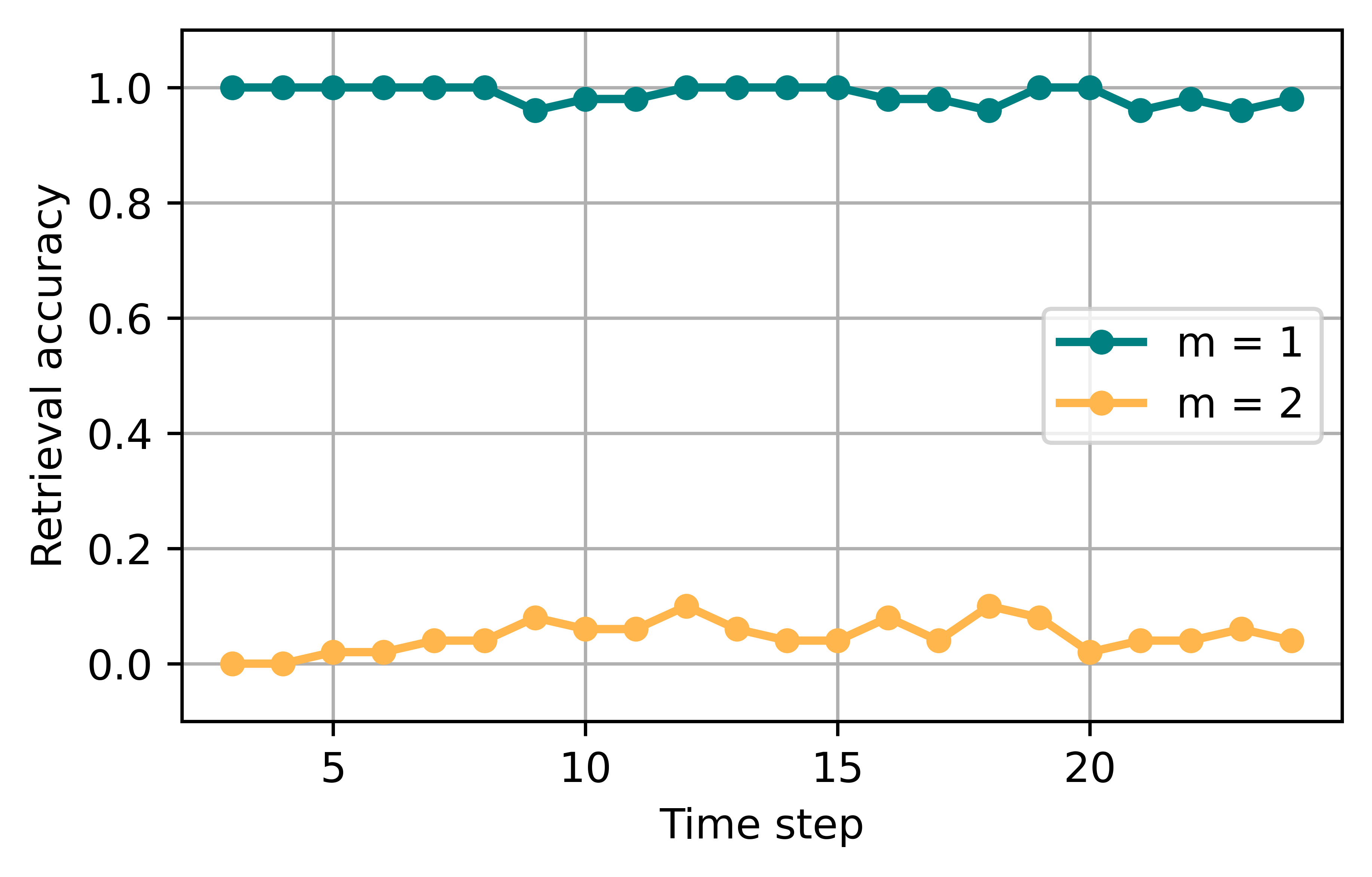}
    \includegraphics[width=0.4\textwidth]{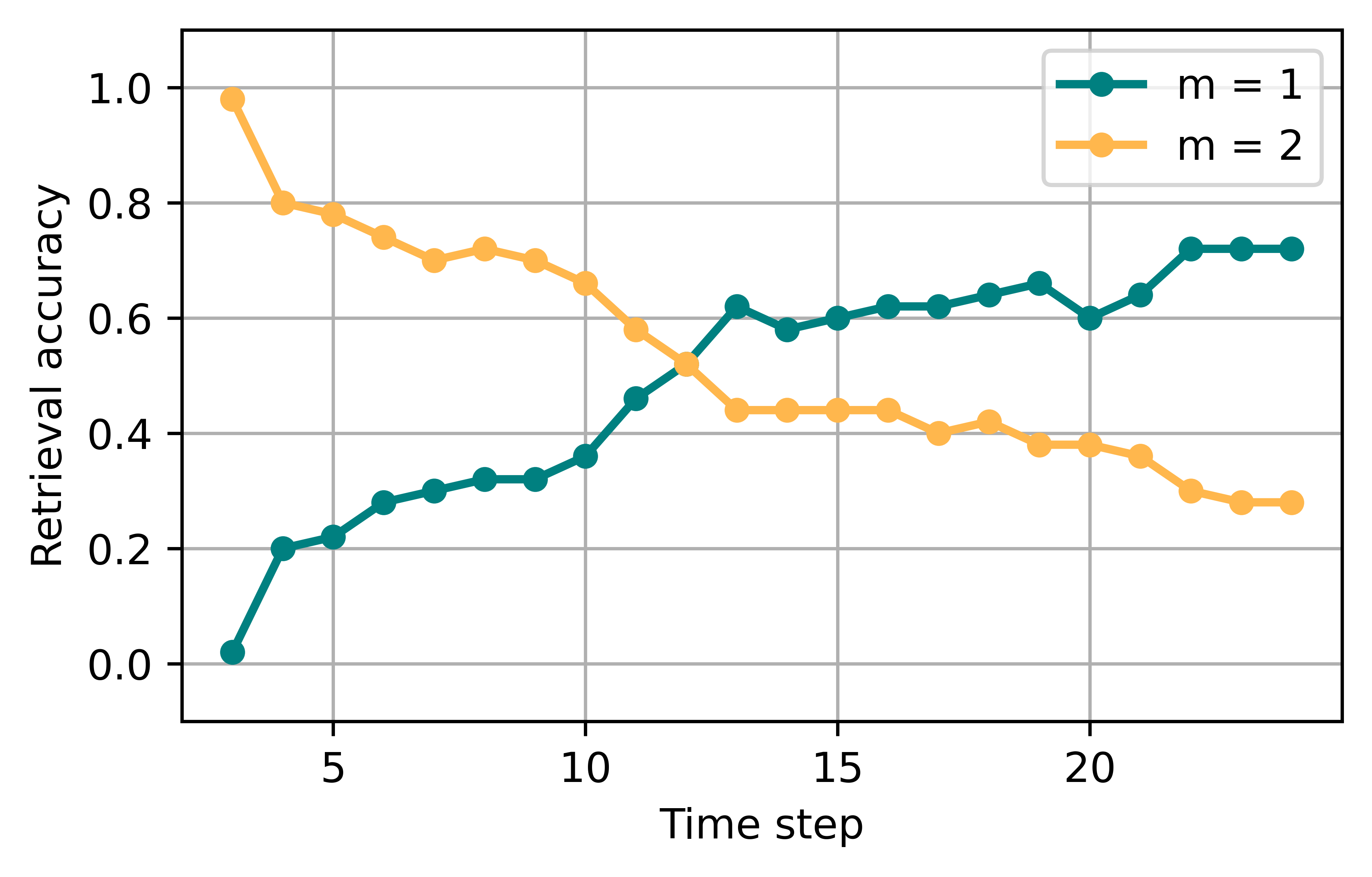}
    \includegraphics[width=0.4\textwidth]{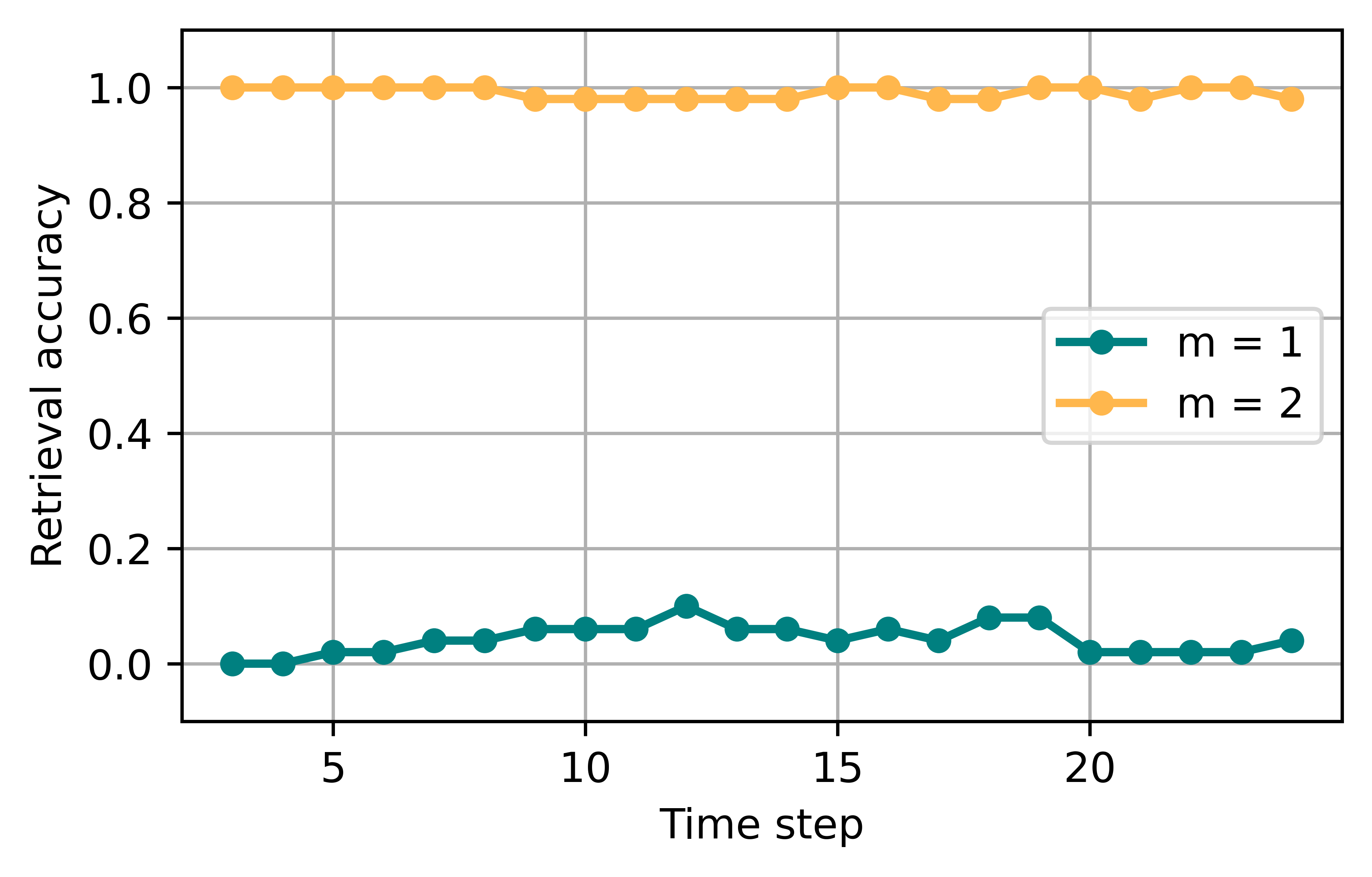}
    \caption{$\mathsf{A}_{2,\cdot}(m,i)$ for $m=1,2$ and $3\leq i\leq 24$. From top to bottom are results for \textsc{Qwen 1.5 14b Chat} (T3), \textsc{Gemma 2 27b Instruct} (T2), and \textsc{Llama 3.1 70b Instruct} (T1).}
    \label{fig:2back-acc-line}
\end{figure}

\begin{figure}[t]
    \centering
    \includegraphics[width=0.4\textwidth]{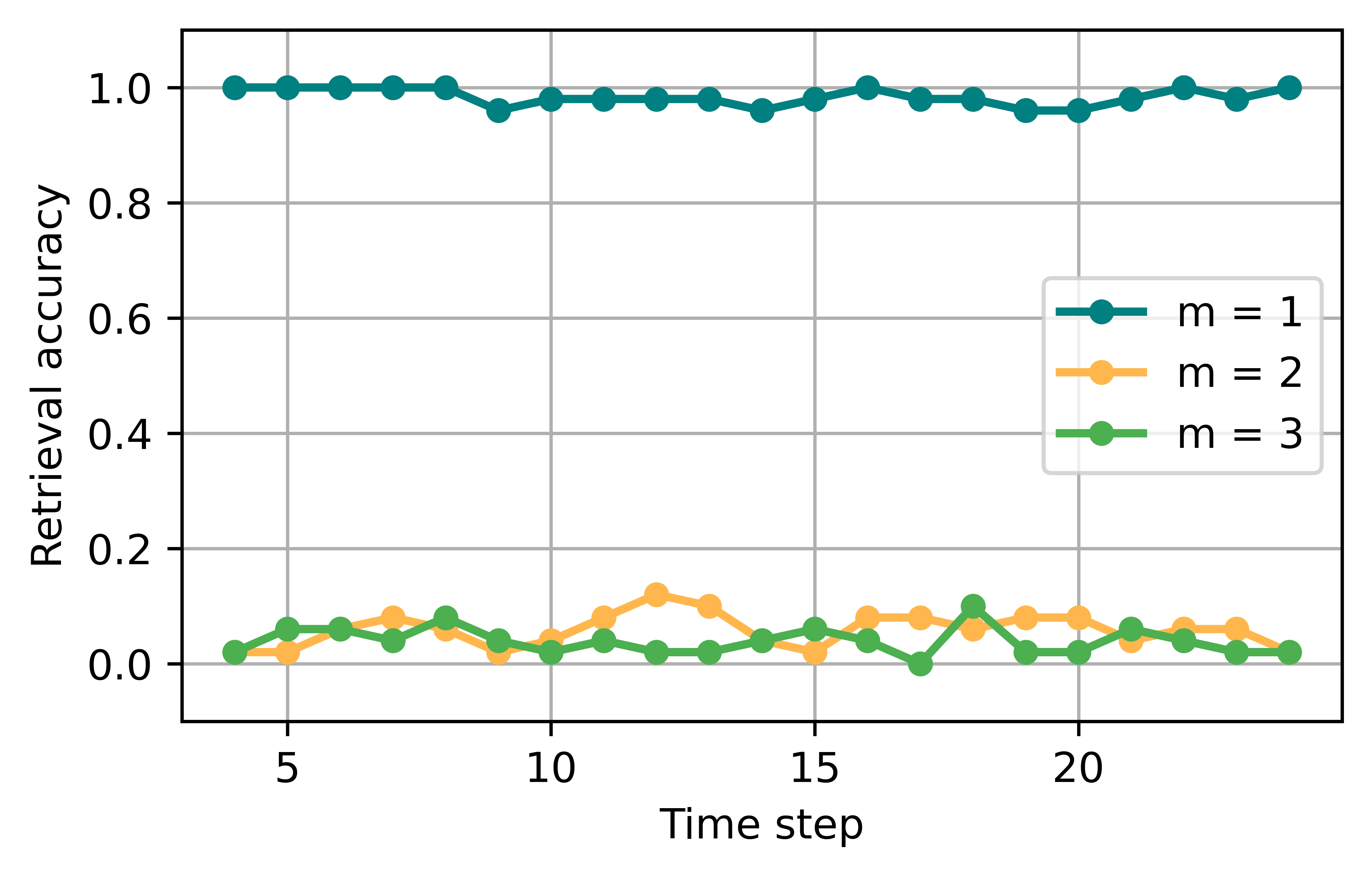}
    \includegraphics[width=0.4\textwidth]{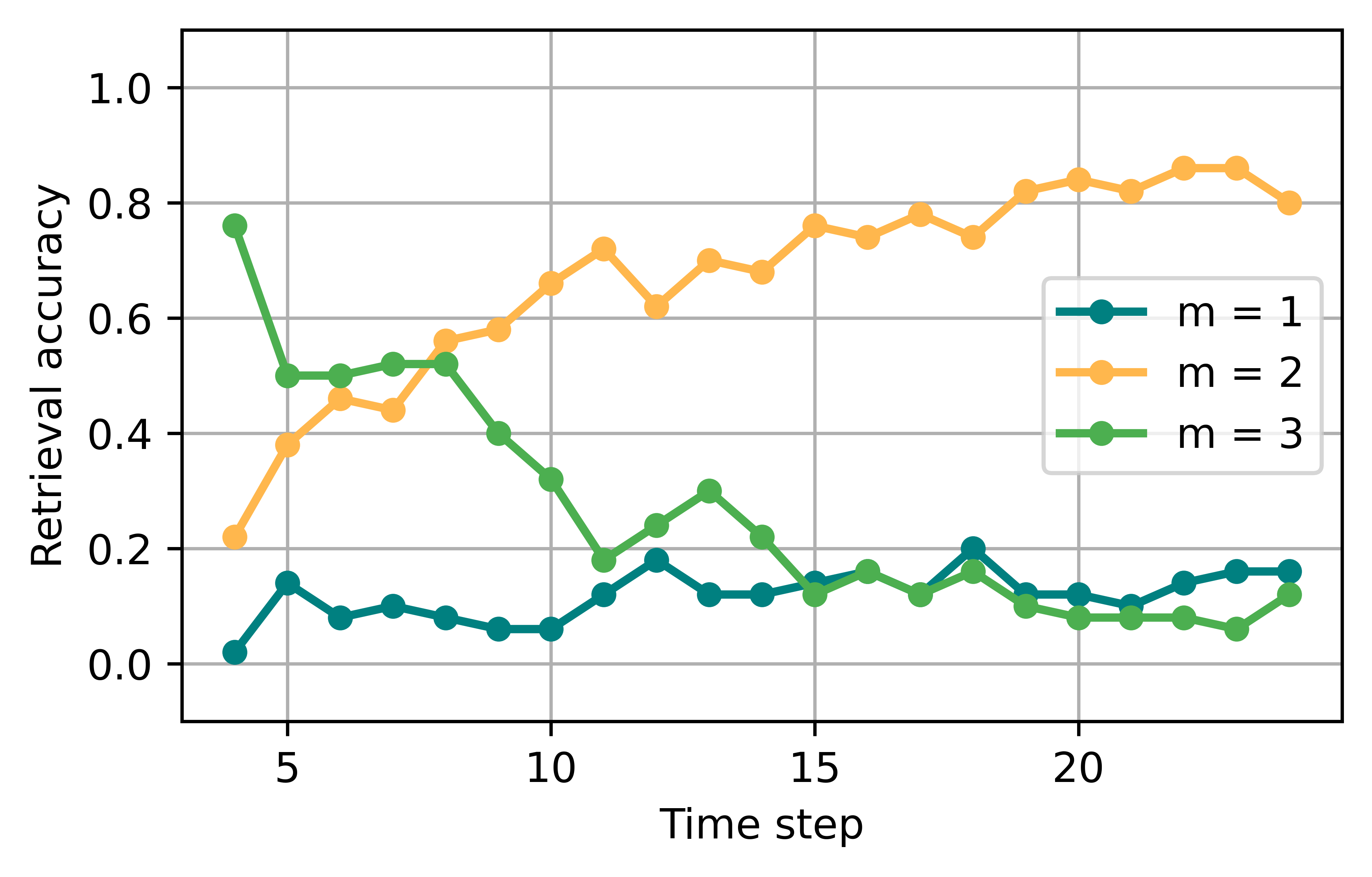}
    \includegraphics[width=0.4\textwidth]{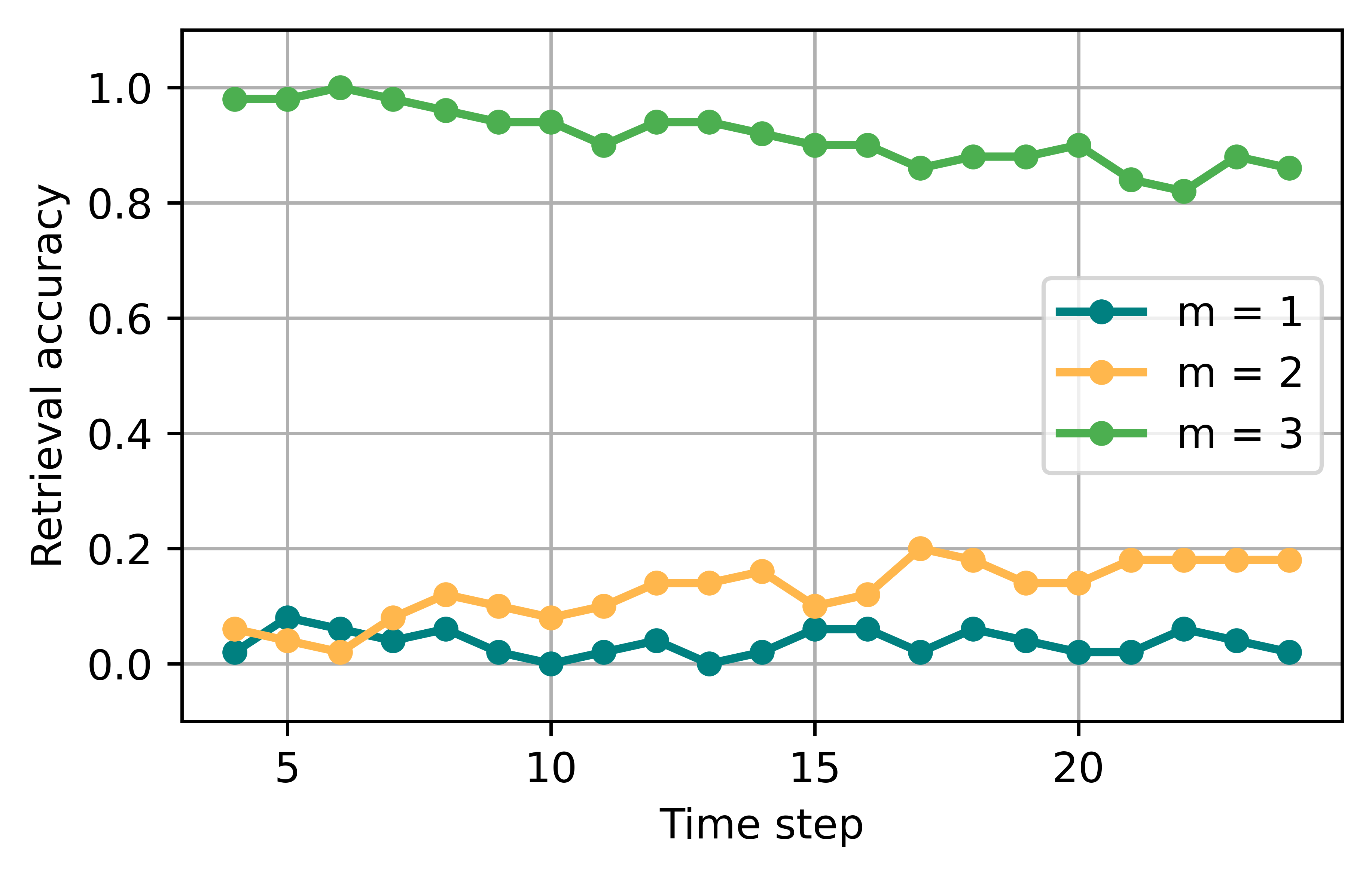}
    \caption{$\mathsf{A}_{3,\cdot}(m,i)$ for $m=1,2,3$ and $4\leq i\leq 24$. From top to bottom are results for \textsc{Qwen 1.5 14b Chat} (T3), \textsc{Gemma 2 27b Instruct} (T2), and \textsc{Llama 3.1 70b Instruct} (T1).}
    \label{fig:3back-acc-line}
\end{figure}

\paragraph{1-back.}
Unsurprisingly, $\mathsf{A}_{1,\cdot}(1,i)$ stays close to 1 for each model as $i$ increases (not shown).

\paragraph{2-back.}
As shown in Figure~\ref{fig:2back-acc-line}:

T3: Throughout the task, $\mathsf{A}_{2,\cdot}(1,i)$ and $\mathsf{A}_{2,\cdot}(2,i)$ stay close to $1$ and $0$, respectively, consistent with findings from Section~\ref{sec:task-comprehension}.

T2: $\mathsf{A}_{2,\cdot}(2,i)$ crosses below $\mathsf{A}_{2,\cdot}(1,i)$ halfway through the task, suggesting a gradual shift from 2-back to 1-back behavior.

T1: Throughout the task, $\mathsf{A}_{2,\cdot}(2,i)$ and $\mathsf{A}_{2,\cdot}(1,i)$ stay close to $1$ and $0$, respectively, contrary to T3.

\paragraph{3-back.}
As shown in Figure~\ref{fig:3back-acc-line}:

T3: Throughout the task, $\mathsf{A}_{3,\cdot}(1,i)$ stays close to $1$ while both $\mathsf{A}_{3,\cdot}(2,i)$ and $\mathsf{A}_{3,\cdot}(3,i)$ stay close to 0, consistent with Section~\ref{sec:task-comprehension}.

T2: After a transient initial lead, $\mathsf{A}_{3,\cdot}(3,i)$ is quickly surpassed by $\mathsf{A}_{3,\cdot}(2,i)$, suggesting yet greater difficulty with task set maintenance.

T1: Throughout the task, $\mathsf{A}_{3,\cdot}(3,i)$ remains close to 1, though it shows a gradual decline over time. Meanwhile, $\mathsf{A}_{3,\cdot}(1,i)$ and $\mathsf{A}_{3,\cdot}(2,i)$ remain low. 

\begin{figure}[t]
  \centering
  \includegraphics[width=0.4\textwidth]{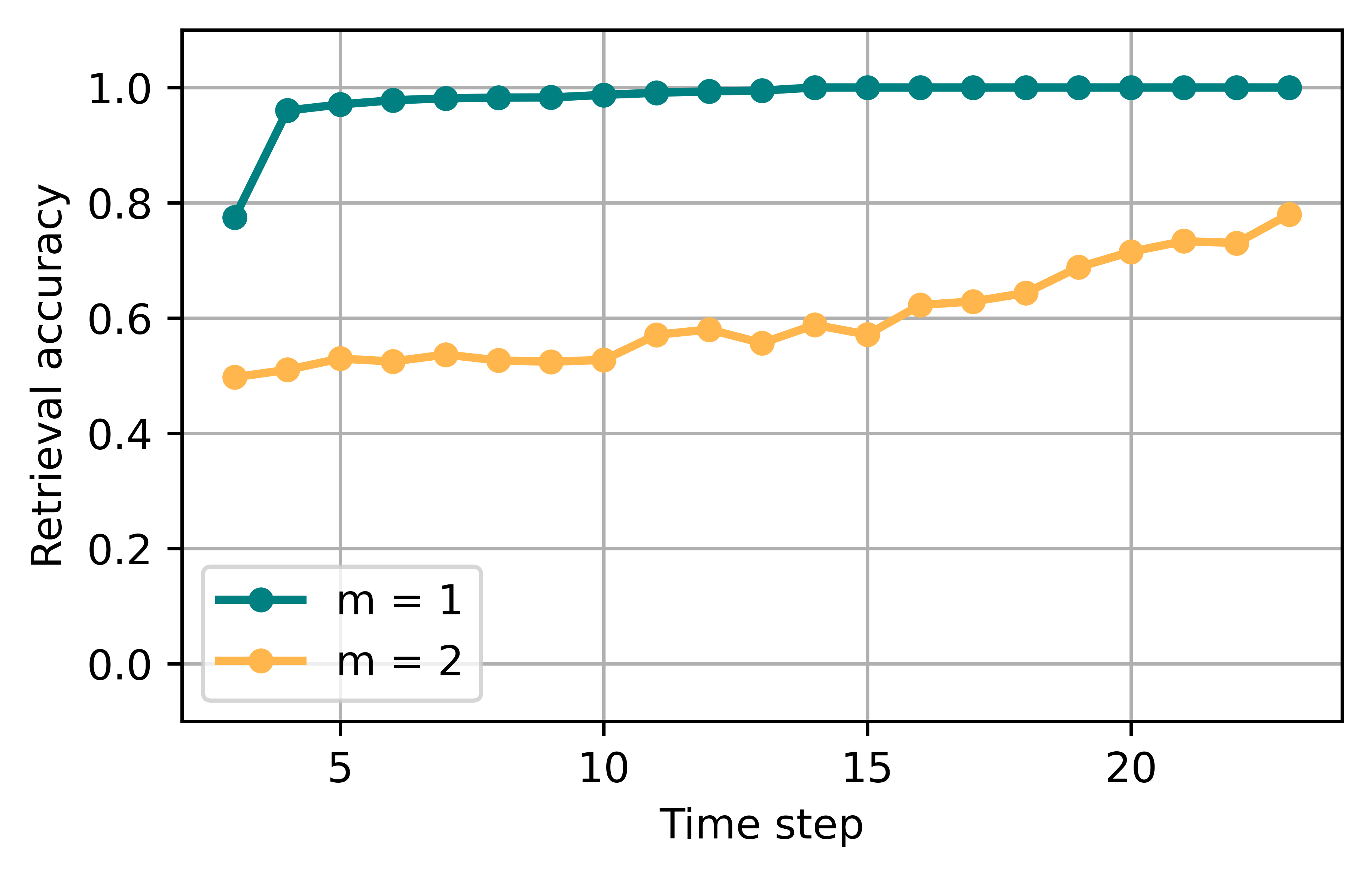}
  \includegraphics[width=0.4\textwidth]{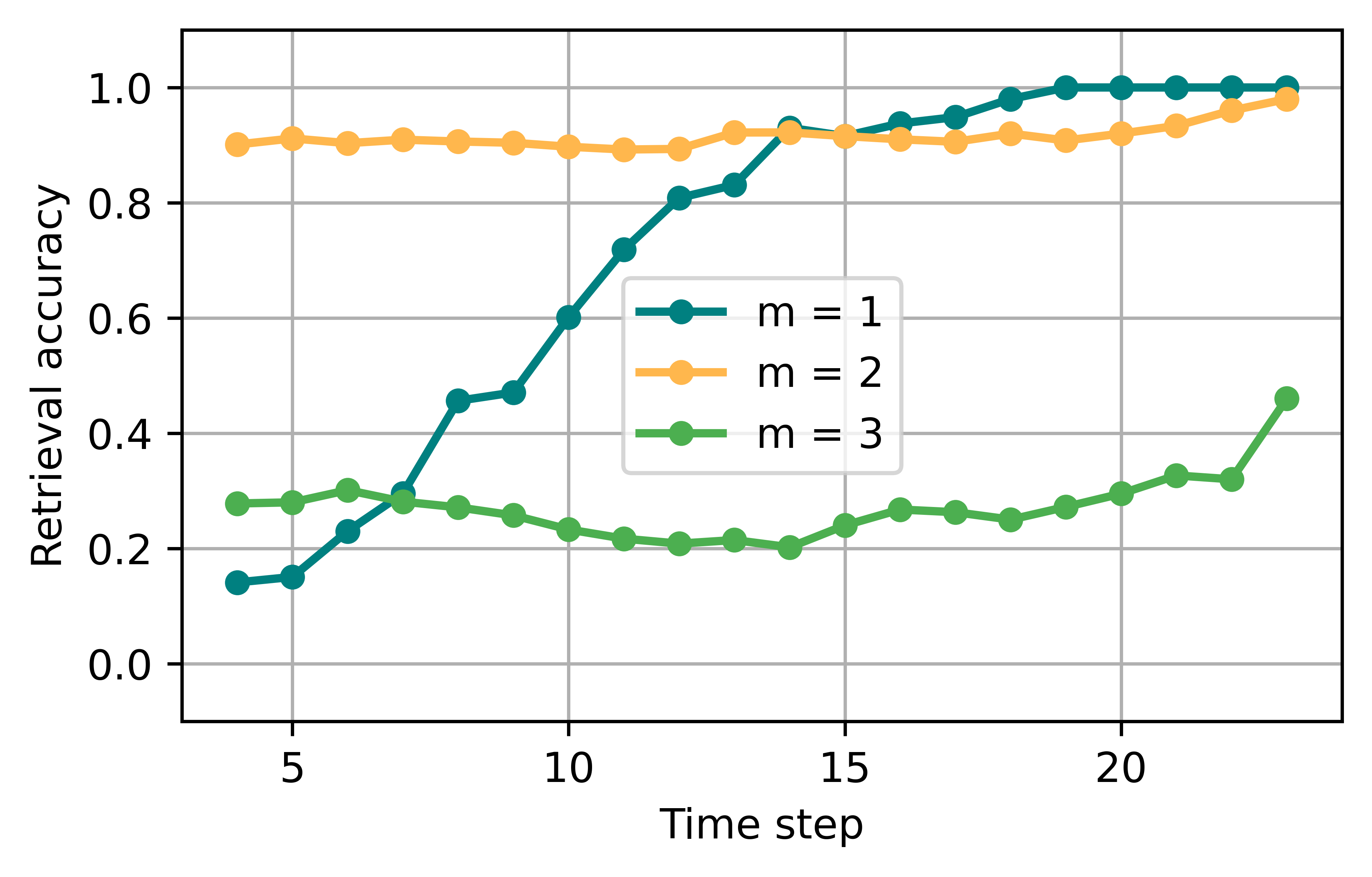}
  \caption{Top: $\mathsf{A}_{2,m}(m,i+1:24)$ for $m=1,2$ and $3\leq i\leq 23$, using \textsc{Gemma 2 27b Instruct} (T2). Bottom: $\mathsf{A}_{3,m}(m,i+1:24)$ for $m=1,2,3$ and $4\leq i\leq 23$, using the same model.}
  \label{fig:error-accumulation}
\end{figure}

\begin{figure}[t]
  \centering
  \includegraphics[width=0.9\columnwidth]{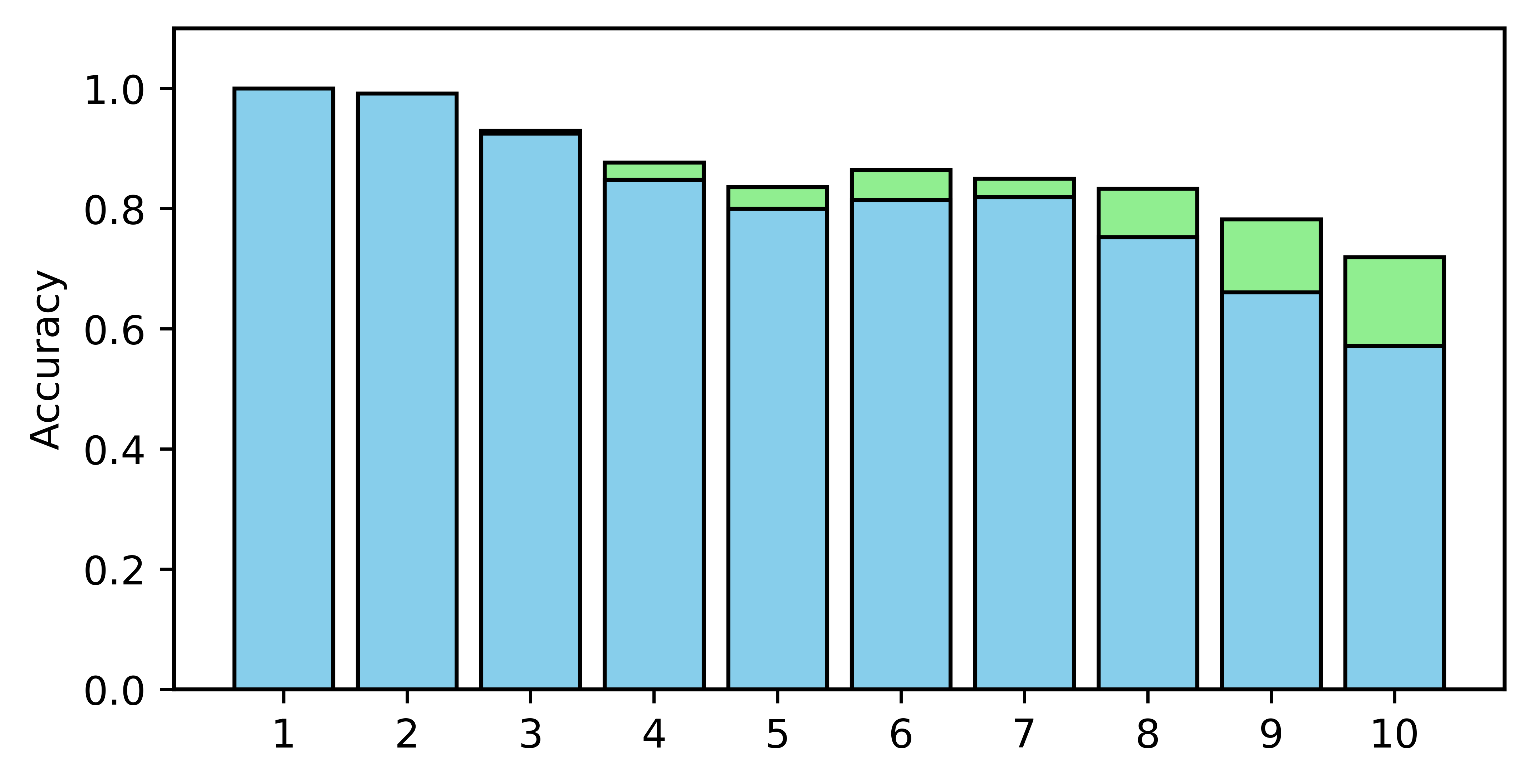}
  \includegraphics[width=0.9\columnwidth]{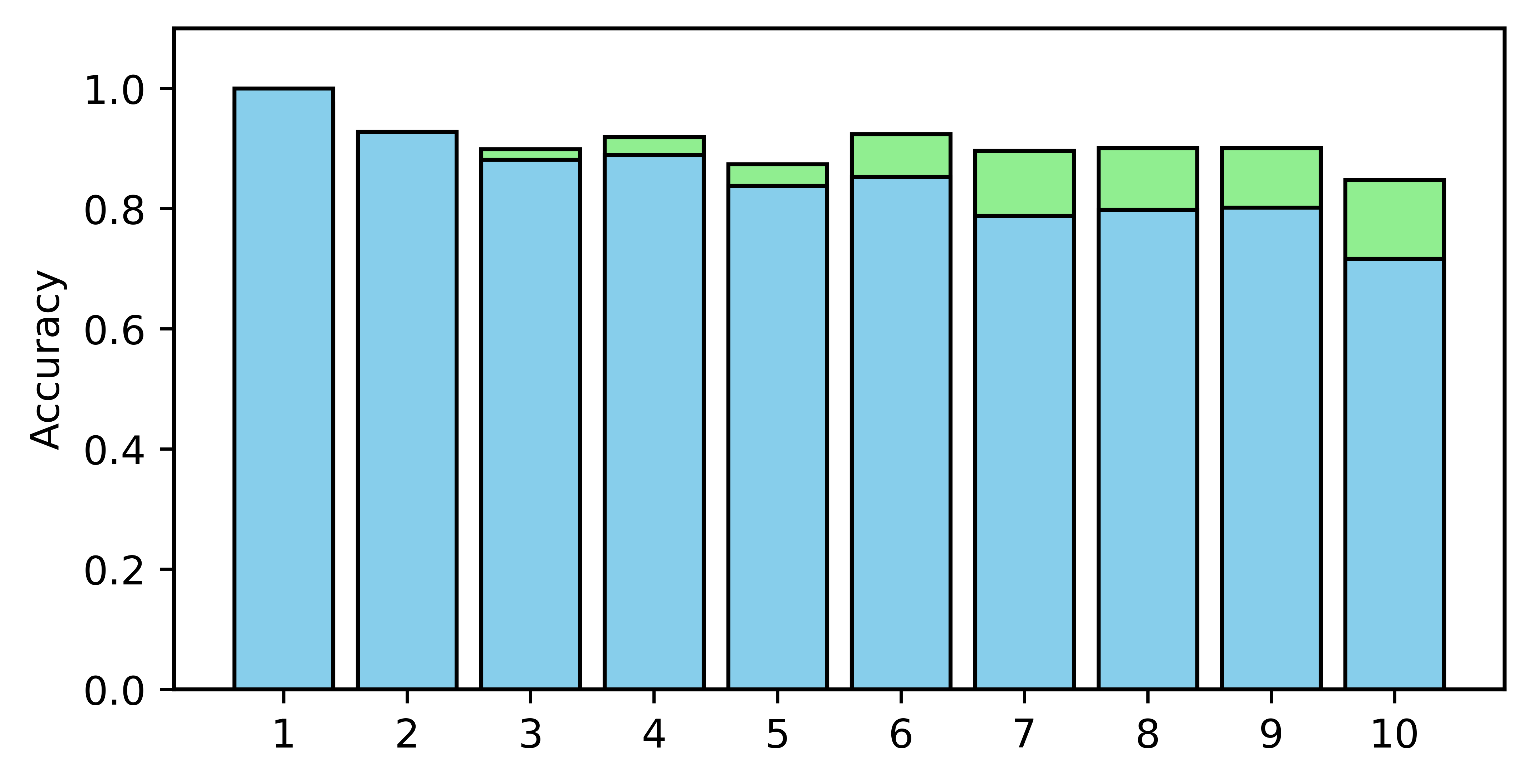}
  \caption{1-back to 10-back accuracies for \textsc{Llama 3.1 70b Instruct} with (bottom) and without (top) curriculum learning. Each full bar corresponds to task (identical/different categorization) accuracy. The blue portion corresponds to retrieval accuracy.}
  \label{fig:10back-acc}
\end{figure}

\begin{figure}[t]
  \centering
  \begin{minipage}{0.24\textwidth}
    \centering
    \includegraphics[width=\linewidth]{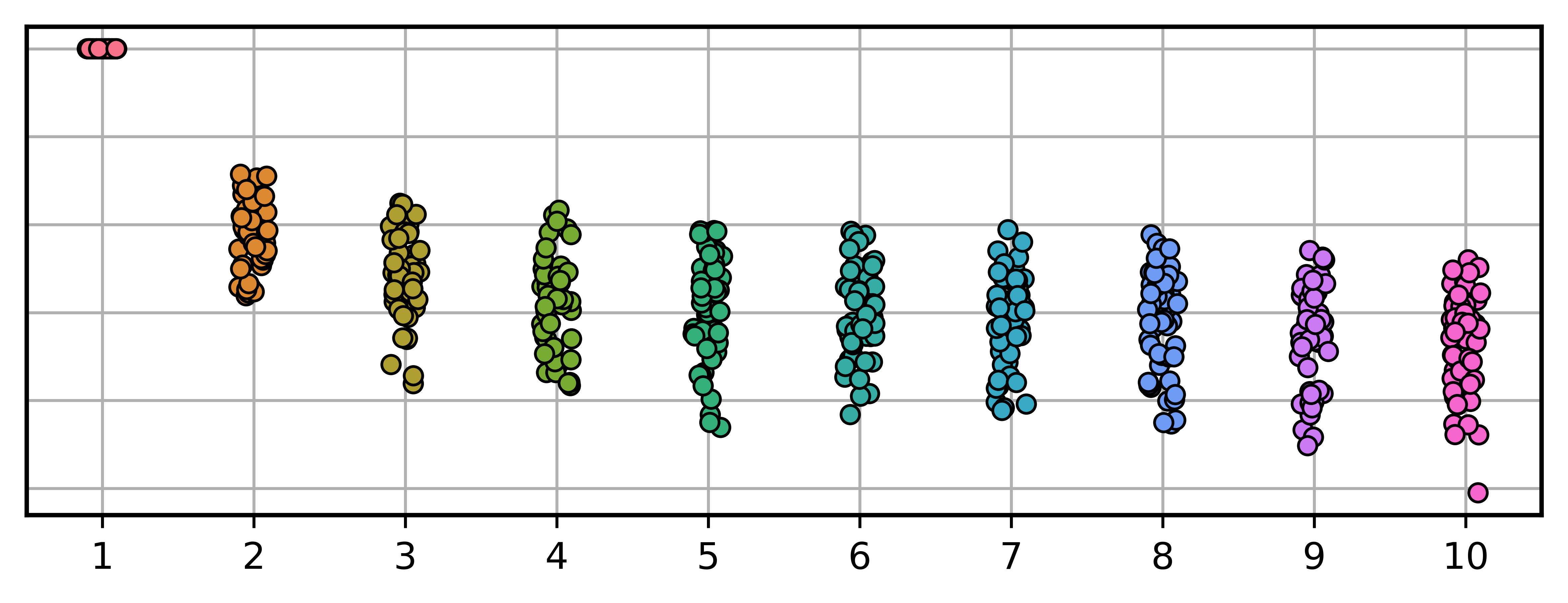}
  \end{minipage}\hfill
  \begin{minipage}{0.24\textwidth}
    \centering
    \includegraphics[width=\linewidth]{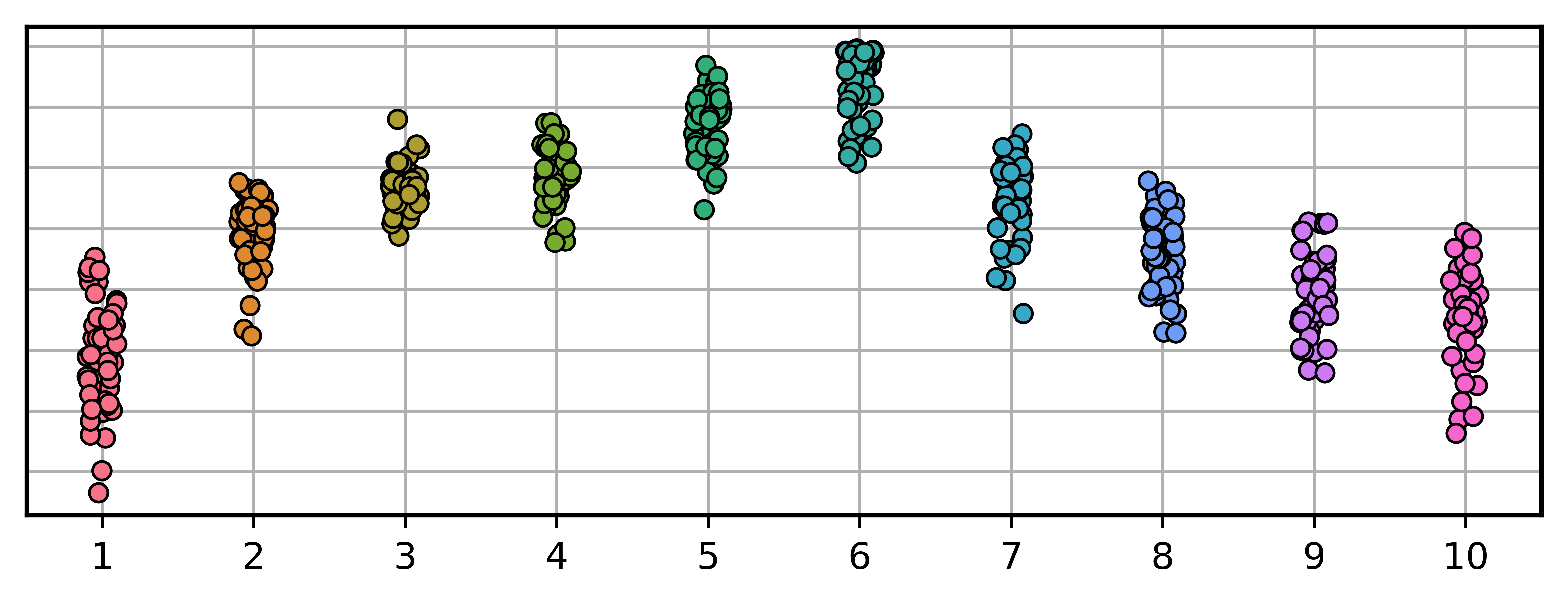}
  \end{minipage}

  \vspace{0.1cm}
  
  \begin{minipage}{0.24\textwidth}
    \centering
    \includegraphics[width=\linewidth]{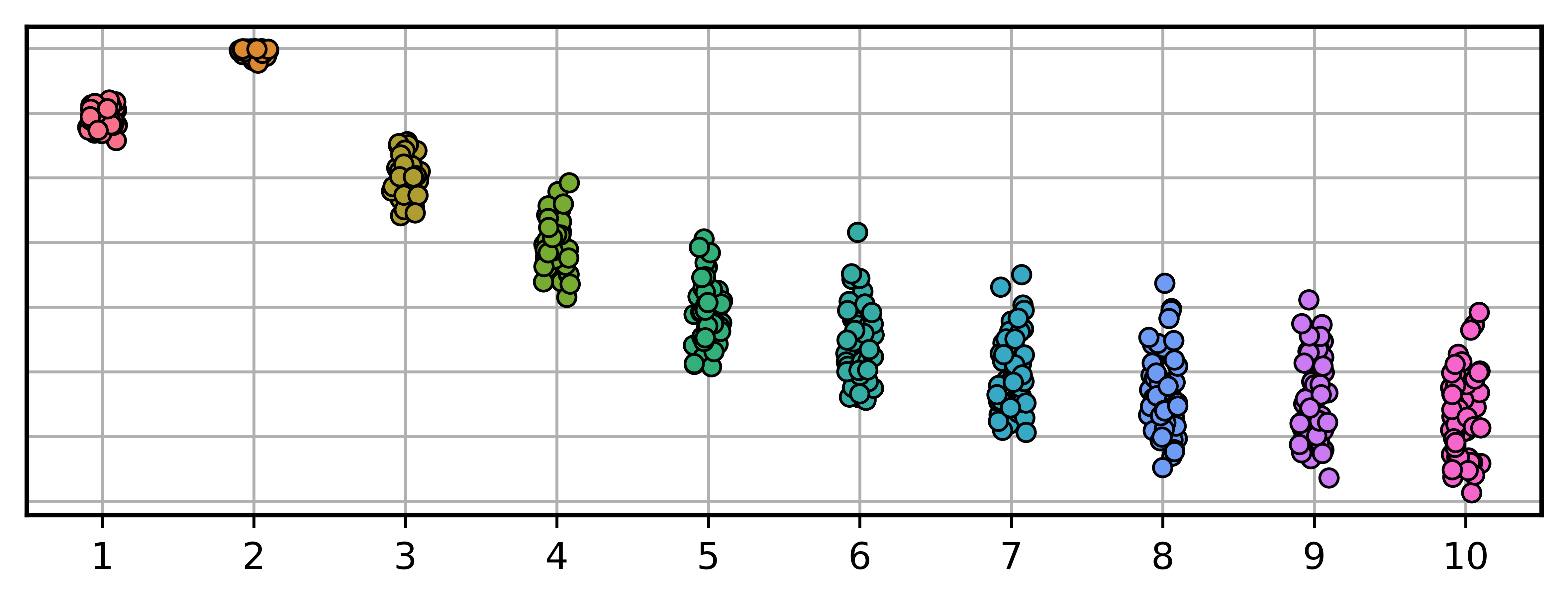}
  \end{minipage}\hfill
  \begin{minipage}{0.24\textwidth}
    \centering
    \includegraphics[width=\linewidth]{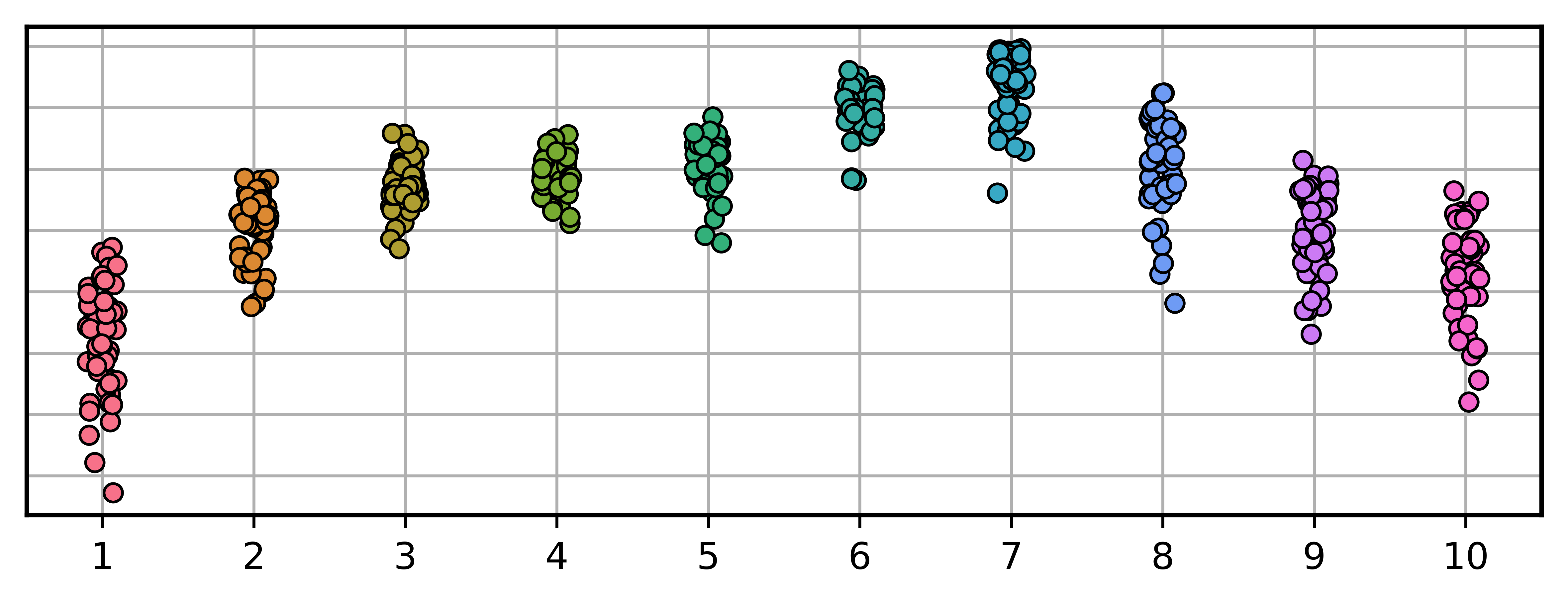}
  \end{minipage}

  \vspace{0.1cm}
  
  \begin{minipage}{0.24\textwidth}
    \centering
    \includegraphics[width=\linewidth]{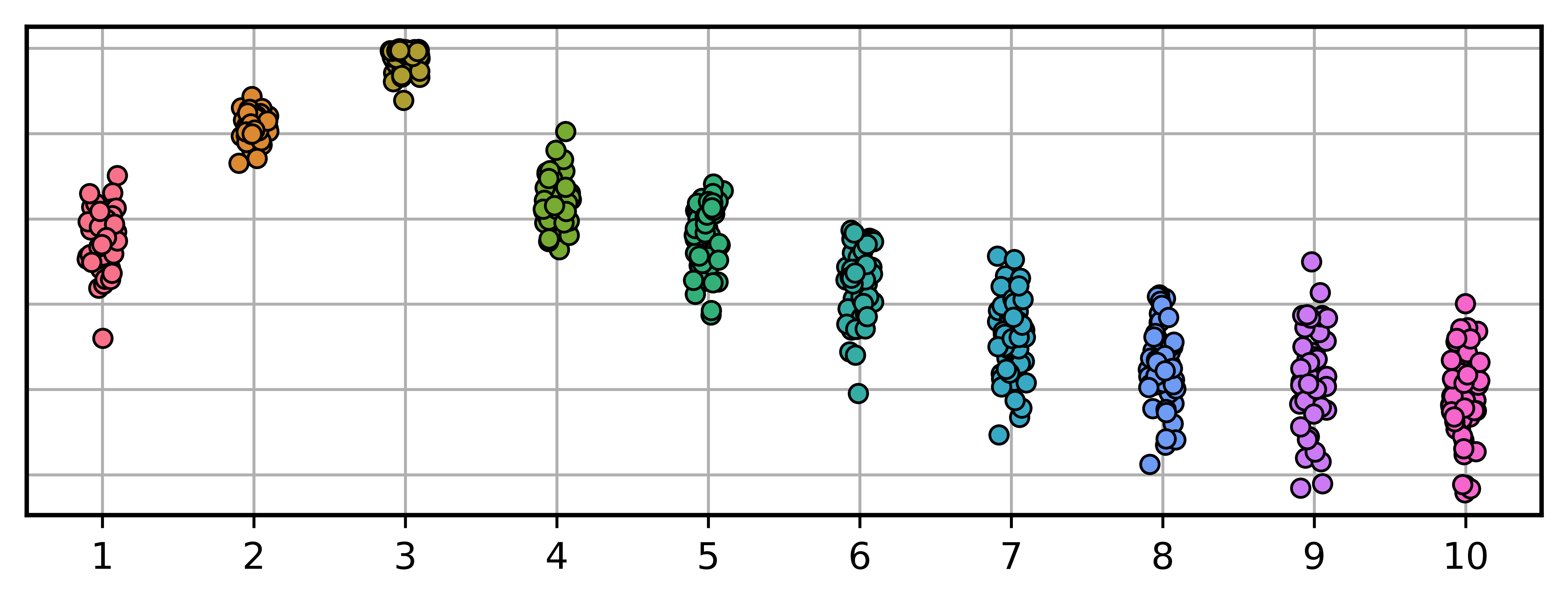}
  \end{minipage}\hfill
  \begin{minipage}{0.24\textwidth}
    \centering
    \includegraphics[width=\linewidth]{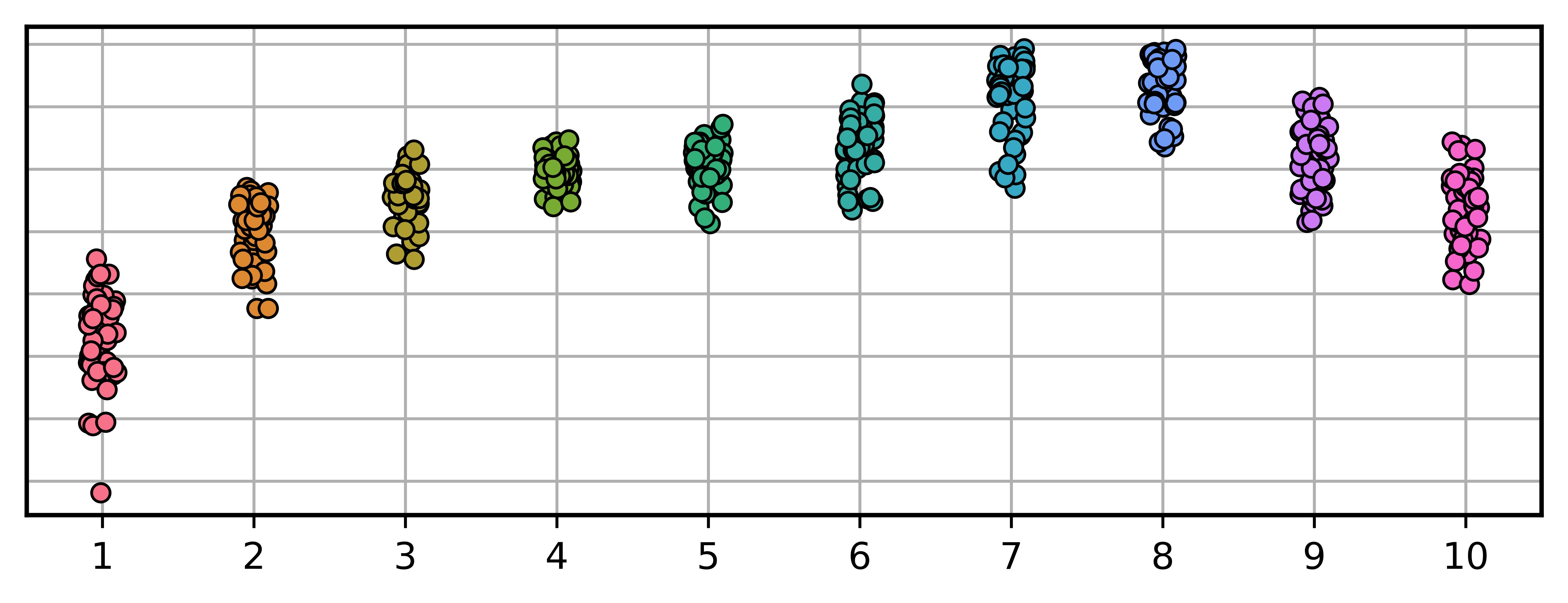}
  \end{minipage}

  \vspace{0.1cm}
  
  \begin{minipage}{0.24\textwidth}
    \centering
    \includegraphics[width=\linewidth]{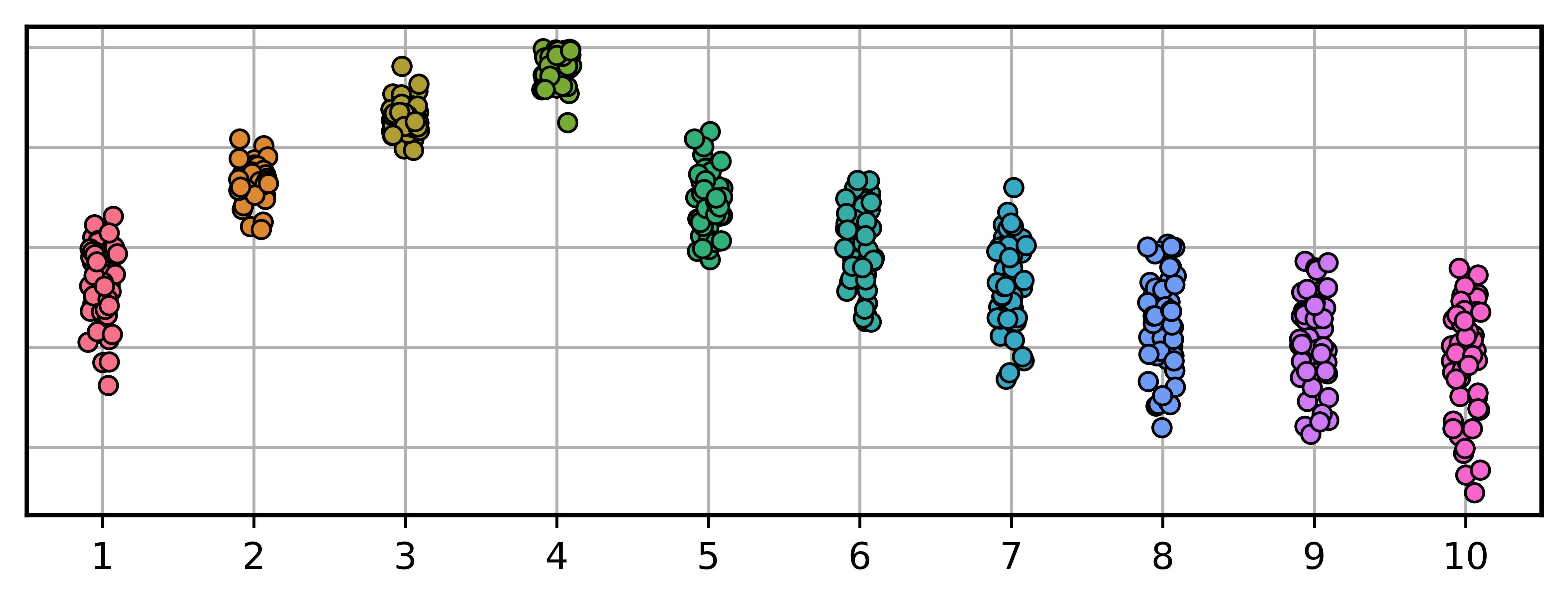}
  \end{minipage}\hfill
  \begin{minipage}{0.24\textwidth}
    \centering
    \includegraphics[width=\linewidth]{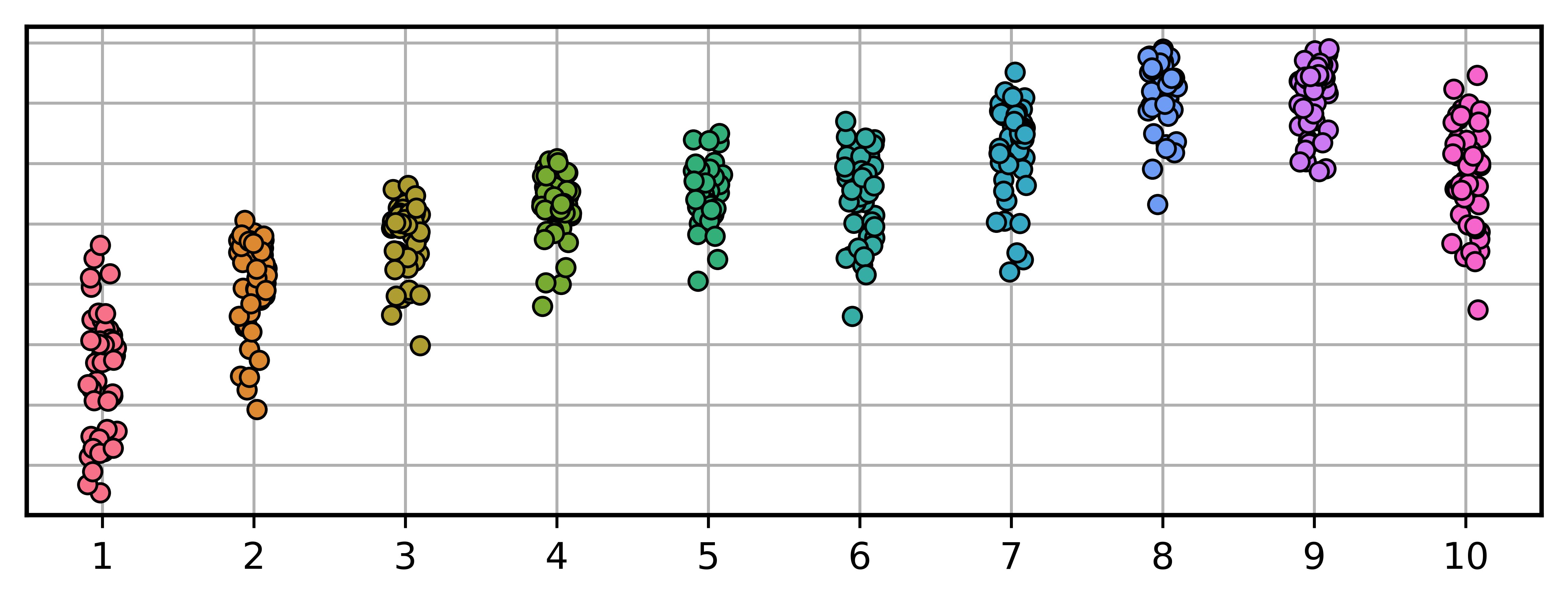}
  \end{minipage}

  \vspace{0.1cm}
  
  \begin{minipage}{0.24\textwidth}
    \centering
    \includegraphics[width=\linewidth]{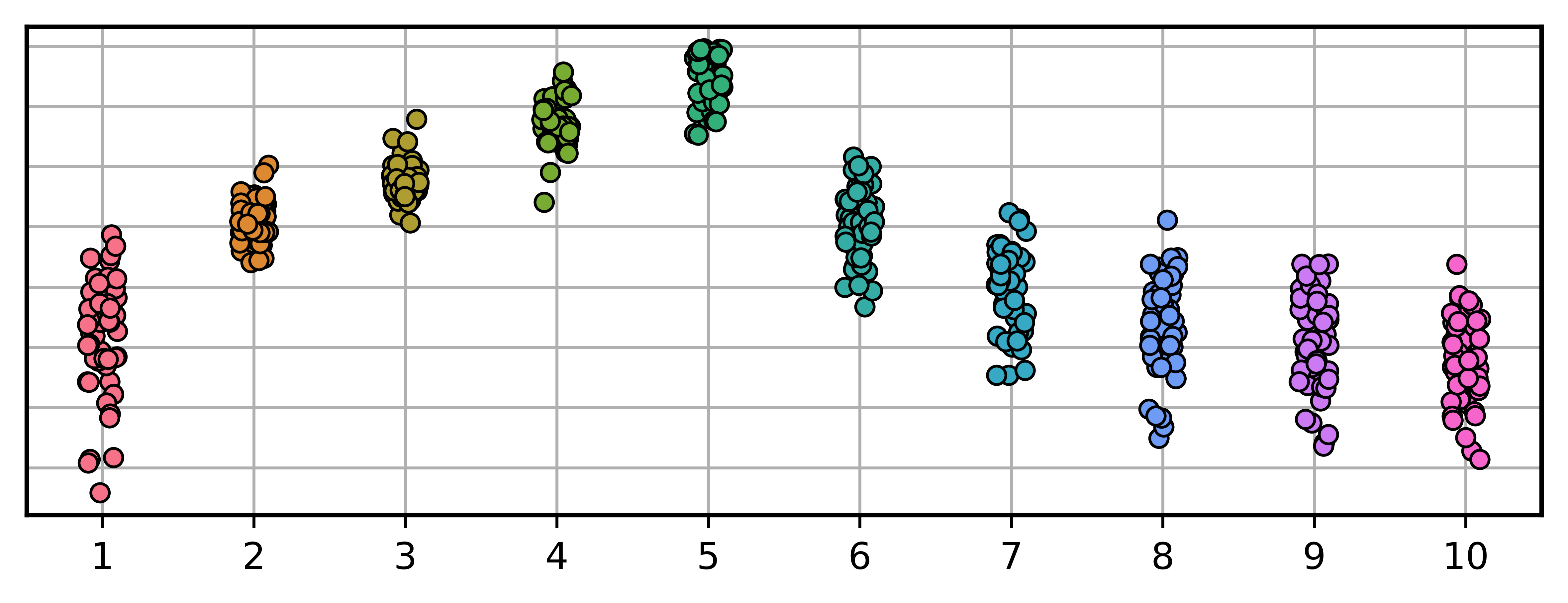}
  \end{minipage}\hfill
  \begin{minipage}{0.24\textwidth}
    \centering
    \includegraphics[width=\linewidth]{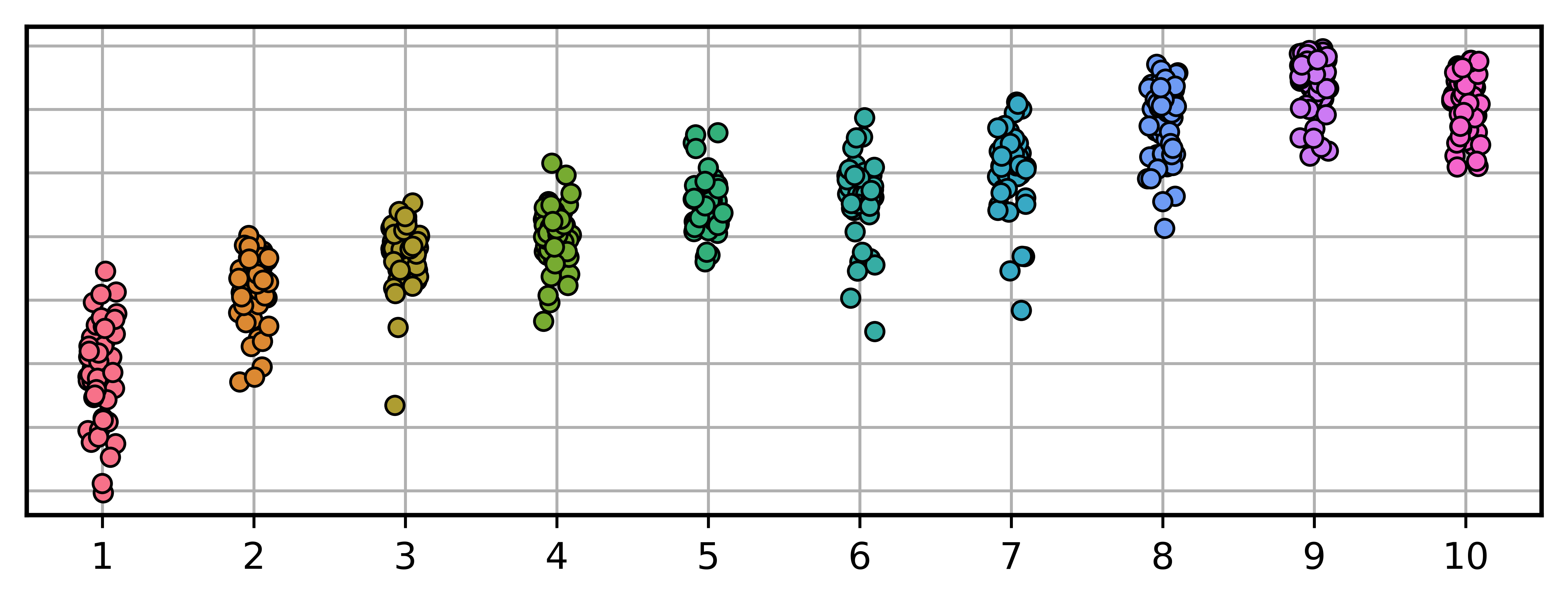}
  \end{minipage}
  
  \caption{Retrieval log probabilities for 1-back to 10-back task continuations under 1-back to 10-back task instructions for \textsc{Llama 3.1 70b Instruct} (T1).}
  \label{fig:10back-strip}
\end{figure}

\paragraph{Effect of error accumulation.}
Despite 2-back instructions and demonstrations, the T2 model gradually drifts toward 1-back consistent responses over time, suggesting that the accumulation of 1-back consistent errors may have significantly biased subsequent responses. To test this hypothesis, we manipulate the model's response history by providing $m$-back consistent responses for \textit{i} steps following $n$-back instructions and demonstrations. We then compute the average $m$-back accuracy for time steps $i+1$ through 24, denoted as $\mathsf{A}_{n,m}(m,i+1:24)$. Figure~\ref{fig:error-accumulation} shows that, as 1-back errors accumulate, 1-back responses are increasingly favored by the T2 model for subsequent steps, despite 2- or 3-back instructions and demonstrations. In comparison, both $\mathsf{A}_{2,2}(2,i+1:24)$ and $\mathsf{A}_{3,3}(3,i+1:24)$ remain relatively low, showing that correct responses do not bias subsequent answers to the same degree.

\subsection{T1 Model Performance as \textit{N} Increases}
\label{sec:llama-70b-10bck}

Given that the best model, \textsc{Llama-3.1-70b-Instruct}, performs well for 1 through 3-back tasks, we would like to know how its performance might change for larger $n$'s. Figure~\ref{fig:10back-acc} shows that the retrieval accuracy gradually declines as $n$ increases; although, even at $n=8,9,10,$ the model is still able to exactly retrieve the correct letters 75.25\%, 66.08\%, and 57.1\% of the time, which translates to task accuracies of 83.33\%, 78.25\%, and 71.92\%. In addition, we measure $\mathsf{P}_{n,m}$ for each $n,m\in\{1,2,3,...,10\}$, as shown in Figure~\ref{fig:10back-strip}. We notice that $\max_{m}\mathsf{P}_{n,m}=\mathsf{P}_{n,n}$ for $1\leq n<10$. Moreover, $\mathsf{P}_{n,m}$ tends to decrease symmetrically as $m$ deviates from $n$. We argue that this pattern points to true $n$-back task understanding.

\subsection{Curriculum Learning}
\label{sec:curriculum-learning}

The practice of training models on examples of increasing difficulty is known in machine learning as \textit{curriculum learning} \citep{bengio2009curriculum}. Here, we repeat the experiments from Section~\ref{sec:llama-70b-10bck} with in-context curriculum learning to gradually familiarize the model with the task. Specifically, before prompting \textsc{Llama 3.1 70b Instruct} to perform an $n$-back task, we provide instructions and demonstrations that include letter sequences and corresponding correct responses for tasks ranging from 1-back to $n$-back. {\color{black}{For example, to prepare the model for the 4-back task, we prepend 4 complete demonstration sequences (1-back to 4-back) to the context, before starting the test sequence.}} As shown in Figure~\ref{fig:10back-acc}, this approach leads to significant improvements in performance for larger $n$ values. The model achieves retrieval accuracies of 79.83\%, 80.17\%, and 71.67\% and task accuracies of 90.08\%, 90.08\%, and 84.75\% for $n=8,9,10$.

\begin{table}
    \centering
    \footnotesize
    \setlength\tabcolsep{5pt}
    \begin{tabular}{c|c|c}
        \toprule
        \textbf{Model} & \textbf{2bk} & \textbf{3bk} \\
        \midrule
        \textsc{Llama 3.1 70b Instr.} & 0.99 (\textendash.00) & 0.62 (\textendash.31) \\
        \textsc{Gemma 2 27b Instr.} & 0.61 (+.04) & 0.31 (\textendash.05) \\
        \textsc{Qwen 1.5 14b Chat} & N/A & N/A \\
        \bottomrule
    \end{tabular}
    \normalsize
    \caption{Retrieval accuracies on 2-back and 3-back tasks, for representative models, with interactive demos.}
    \label{tab:int-demo}
\end{table}

\subsection{Interactive Demo}
\label{sec:interactive-demo}

We explore an alternative prompting strategy that more closely mirrors human study paradigms. After receiving task instructions, human participants typically go through brief demo sequences with an experimenter to confirm their understanding. For 2-back trials, we interleave short example sequences of four letters in the forms A-B-A-C and A-B-C-B. Feedback is given for each model response. If a model provides two consecutive correct answers (retrieval and label) within 10 attempts, we proceed with the test sequence. A similar procedure is applied for 3-back trials.

For both 2-back and 3-back tasks, \textsc{Qwen 1.5 14b Chat} (T3) fails to achieve two consecutive correct answers after 10 demo sequences, further confirming the model's difficulty with task comprehension. {\color{black}{The complete 2-back dialogue is included in Appendix~\ref{appendix:dialogue}.}} Interestingly, \textsc{Gemma 2 27b Instruct} (T2) performs better on 2-back trials compared to the original experiments but does worse on 3-back trials, as shown in Table~\ref{tab:int-demo}.
\textsc{Llama 3.1 70b Instruct} (T1) maintains high performance at 99\% on 2-back trials but shows a significant drop in performance on 3-back.

\begin{figure}[t]
  \centering
  \includegraphics[width=0.9\columnwidth]{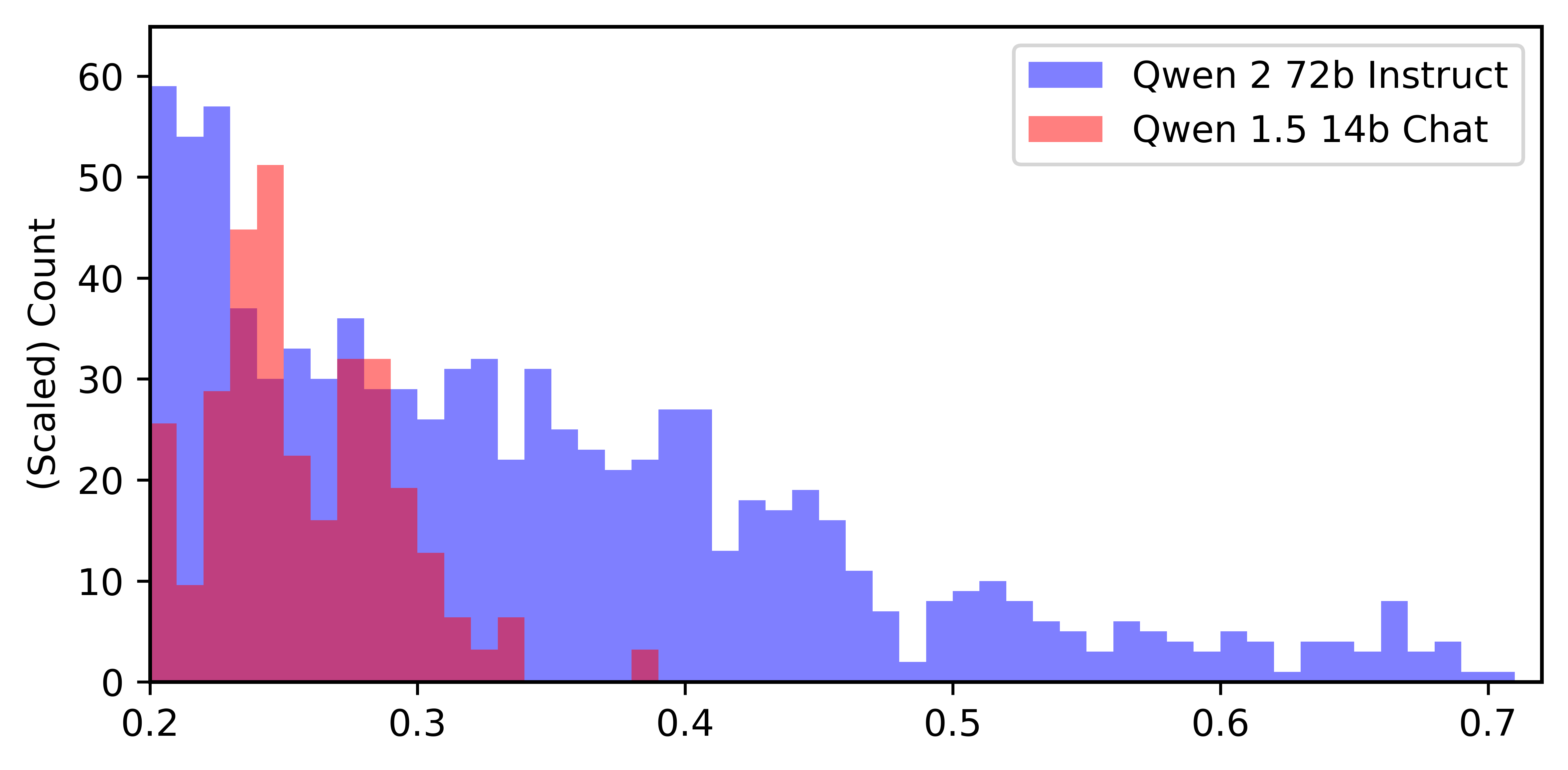}
  \caption{2-back MRAT counts between 0.2 and 1 for \textsc{Qwen 1.5 14B Chat} (T3) and \textsc{Qwen 2 72b Instruct} (T1), aggregated over all layers, heads, and trials. \textsc{Qwen 1.5 14b Chat} counts are scaled by a factor of $\frac{\textsc{Qwen 2 72b}\text{ Attention Count}}{\textsc{Qwen 1.5 14b}\text{ Attention Count}}=3.2$.}
  \label{fig:attn-comp}
\end{figure}

\begin{figure}[ht]
  \centering
  \includegraphics[width=1\columnwidth]{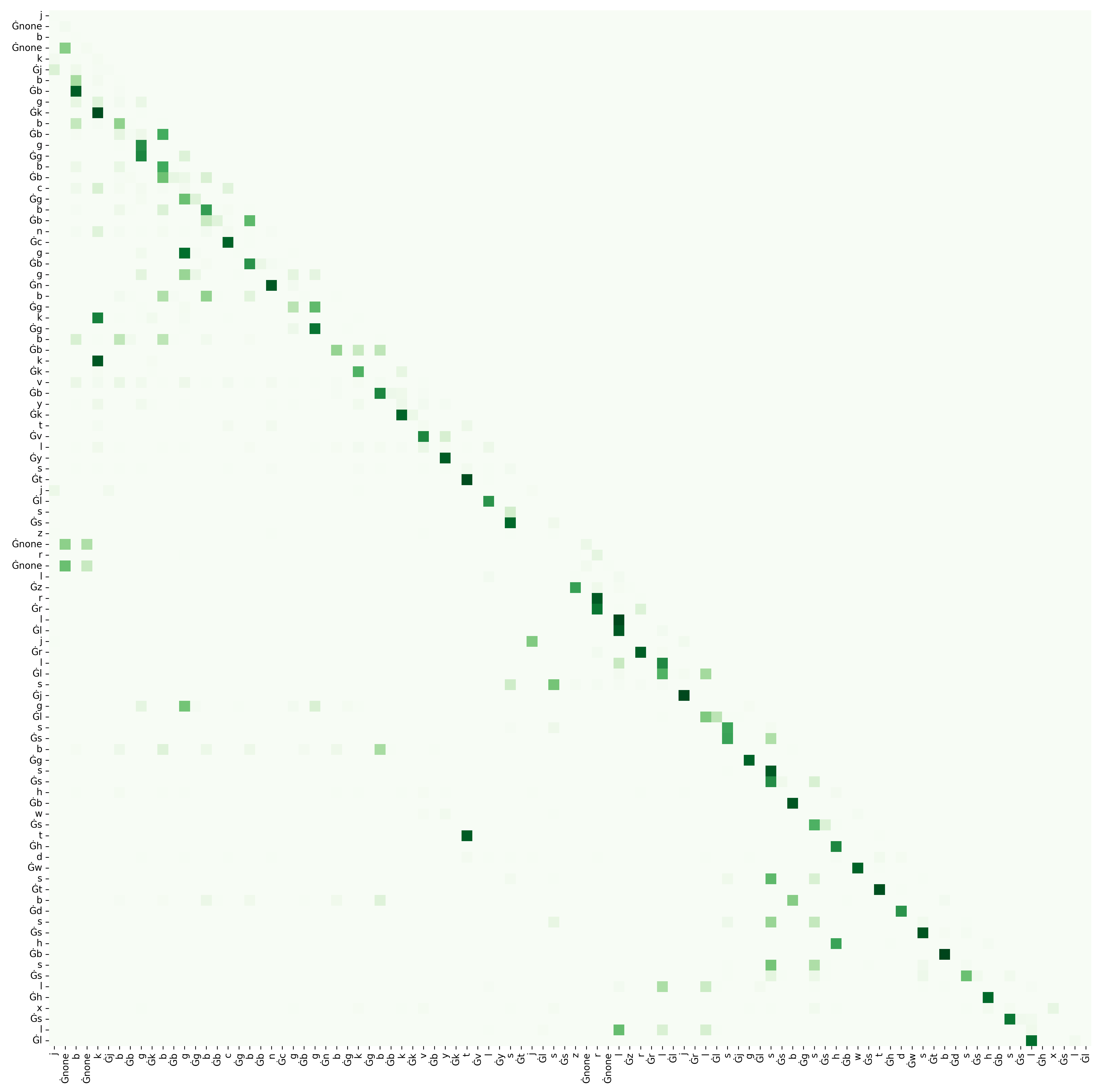}
  \caption{\textsc{Qwen 2 72b Instruct} (T1) attention pattern with the highest MRAT (71.98\%) at trial 48, layer 79, and head 63. The top left and bottom right sections correspond to the demo and test sequences, respectively.}
  \label{fig:attn-map}
\end{figure}

\subsection{Attention Analysis}
\label{sec:attn-analysis}

Attentions in transformer-based language models reveal how much each generated token attends to every preceding token. We hypothesize that, for each retrieval, a more performant model should attend more to the source token from $n$ steps back. This is precisely what we observe in the \textsc{Qwen} models. For each (trial, layer, head), we obtain the \textit{mean retrieval attention} (MRAT) by averaging the attention each retrieved token gives to the correct source token.
Compared to the \textsc{14b} model, \textsc{Qwen 2 72b Instruct} (T1) contains a much larger proportion of high-MRAT attentions (Figure~\ref{fig:attn-comp}), with its highest scoring attention (71.98\%) closely matching our hypothesized pattern (Figure~\ref{fig:attn-map}).
However, \textsc{Llama} models do not exhibit this pattern to the same degree. Attentions in \textsc{Llama 3.1} models are much more diffuse. The maximum MRATs for \textsc{Llama 3.1 8b Instruct} and \textsc{Llama 3.1 70b Instruct} are 4.86\% and 8.52\%, respectively.

\section{Conclusion}
\label{sec:conclusions}

In this work, we apply the $n$-back task, a common working memory test, to a range of language models, identifying three distinct performance tiers.
We find that these tiers differ not only in retrieval accuracy but also in our measure of task understanding and task set maintenance, suggesting that the performance gap is due at least in part to these differences.
We challenge the best model to perform 1 through 10-back tasks, noticing a signature of task comprehension and the benefit of in-context curriculum learning for larger $n$'s.
We find that interactive demos, though closer to human study paradigms, are less effective at conveying the task.
Finally, we notice that more performant models tend to have higher retrieval attentions.

\section{Limitations}
\label{sec:limitations}

\paragraph{Prompt selection.}
Despite our careful selection of prompts and experimentation with various prompting strategies, the potential for more effective prompts or techniques to enhance task understanding remains.

\paragraph{Mechanistic understanding.}
Another limitation is that we do not examine the internal model circuits that may be responsible for inferring and maintaining task sets. However, our experiments with the $n$-back paradigm provide a good starting point for future research. Causal interventions on smaller models may yield insights into the underlying mechanisms.

\paragraph{\textsc{Llama 3.1} model attentions.} 
As mentioned in Section~\ref{sec:attn-analysis}, \textsc{Llama} models seem to have much more diffuse attentions. Whereas \textsc{Qwen 1.5 14b Chat} and \textsc{Qwen 2 72b Instruct} reach maximum MRATs of 38.95\% and 71.98\%, respectively, the same values for both \textsc{Llama} models are under 10\%, even though \textsc{Qwen} \citep{bai2023qwen} and \textsc{Llama} \citep{dubey2024llama} models both use \textit{grouped query attention} \citep{ainslie2023gqa}. We are puzzled by this difference and call for closer examination in future work.

\section{Acknowledgments}

We thank the anonymous reviewers for their helpful feedback.
XH would also like to acknowledge the participants in the Brown MICAAD study, who partially inspired this project.

\bibliography{acl}

\clearpage

\appendix


\section{Interactive Demo Dialogue}
\label{appendix:dialogue}

{\color{black}
\textsc{Qwen 1.5 14b Chat} (T3) 2-back interactive demo dialogue.
}

{
\footnotesize
\begin{verbatim}
<|im_start|>system
You are a participant in a cognitive task. You will
be shown a sequence of letters, one at a time. For
each letter, determine if it is the same as the
letter 2 steps before it. Answer in the following
format: '[current letter] and [letter 2 back] are
[identical/different].' Note that, for the first 2
letters, there won't be any letter 2 steps back;
write 'none' for '[letter 2 back]' in this case.
Let's think step by step.
<|im_end|>

<|im_start|>user
For example, given the sequence t, z, h, z,
the answers should be:
t and none are different.
z and none are different.
h and t are different.
z and z are identical.
Now, given the sequence t, z, t, h,
what should the answers be?
<|im_end|>

<|im_start|>assistant
t and none are different.
z and t are different.
t and z are different.
h and t are different.
<|im_end|>

<|im_start|>user
This is incorrect.
The answers should be:
t and none are different.
(There was no letter 2 steps ago.)
z and none are different.
(There was no letter 2 steps ago.)
t and t are identical.
(The letter 2 steps ago was t.)
h and z are different.
(The letter 2 steps ago was z.)
Now, given the sequence v, n, c, n,
what should the answers be?
<|im_end|>

<|im_start|>assistant
v and none are different.
n and v are different.
c and n are different.
n and n are identical.
<|im_end|>

<|im_start|>user
This is incorrect.
The answers should be:
v and none are different.
(There was no letter 2 steps ago.)
n and none are different.
(There was no letter 2 steps ago.)
c and v are different.
(The letter 2 steps ago was v.)
n and n are identical.
(The letter 2 steps ago was n.)
Now, given the sequence v, n, v, c,
what should the answers be?
<|im_end|>

<|im_start|>assistant
v and none are different.
n and v are different.
v and n are different.
c and v are different.
(The letter 2 steps ago was v.)
<|im_end|>

<|im_start|>user
This is incorrect.
The answers should be:
v and none are different.
(There was no letter 2 steps ago.)
n and none are different.
(There was no letter 2 steps ago.)
v and v are identical.
(The letter 2 steps ago was v.)
c and n are different.
(The letter 2 steps ago was n.)
Now, given the sequence k, l, b, l,
what should the answers be?
<|im_end|>

<|im_start|>assistant
k and none are different.
l and k are different.
b and l are different.
l and l are identical.
<|im_end|>

<|im_start|>user
This is incorrect.
The answers should be:
k and none are different.
(There was no letter 2 steps ago.)
l and none are different.
(There was no letter 2 steps ago.)
b and k are different.
(The letter 2 steps ago was k.)
l and l are identical.
(The letter 2 steps ago was l.)
Now, given the sequence k, l, k, b,
what should the answers be?
<|im_end|>

<|im_start|>assistant
k and none are different.
l and k are different.
k and l are different.
b and k are different.
(The letter 2 steps ago was k.)
<|im_end|>

<|im_start|>user
This is incorrect.
The answers should be:
k and none are different.
(There was no letter 2 steps ago.)
l and none are different.
(There was no letter 2 steps ago.)
k and k are identical.
(The letter 2 steps ago was k.)
b and l are different.
(The letter 2 steps ago was l.)
Now, given the sequence r, f, m, f,
what should the answers be?
<|im_end|>

<|im_start|>assistant
r and none are different.
f and r are different.
m and f are different.
f and f are identical.
<|im_end|>

<|im_start|>user
This is incorrect.
The answers should be:
r and none are different.
(There was no letter 2 steps ago.)
f and none are different.
(There was no letter 2 steps ago.)
m and r are different.
(The letter 2 steps ago was r.)
f and f are identical.
(The letter 2 steps ago was f.)
Now, given the sequence r, f, r, m,
what should the answers be?
<|im_end|>

<|im_start|>assistant
r and none are different.
f and r are different.
r and f are different.
m and r are different.
(The letter 2 steps ago was r.)
<|im_end|>

<|im_start|>user
This is incorrect.
The answers should be:
r and none are different.
(There was no letter 2 steps ago.)
f and none are different.
(There was no letter 2 steps ago.)
r and r are identical.
(The letter 2 steps ago was r.)
m and f are different.
(The letter 2 steps ago was f.)
Now, given the sequence y, p, w, p,
what should the answers be?
<|im_end|>

<|im_start|>assistant
y and none are different.
p and y are different.
w and p are different.
p and p are identical.
<|im_end|>

<|im_start|>user
This is incorrect.
The answers should be:
y and none are different.
(There was no letter 2 steps ago.)
p and none are different.
(There was no letter 2 steps ago.)
w and y are different.
(The letter 2 steps ago was y.)
p and p are identical.
(The letter 2 steps ago was p.)
Now, given the sequence y, p, y, w,
what should the answers be?
<|im_end|>

<|im_start|>assistant
y and none are different.
p and y are different.
y and p are different.
w and y are different.
(The letter 2 steps ago was y.)
<|im_end|>

<|im_start|>user
This is incorrect.
The answers should be:
y and none are different.
(There was no letter 2 steps ago.)
p and none are different.
(There was no letter 2 steps ago.)
y and y are identical.
(The letter 2 steps ago was y.)
w and p are different.
(The letter 2 steps ago was p.)
Now, given the sequence s, j, g, j,
what should the answers be?
<|im_end|>

<|im_start|>assistant
s and none are different.
j and s are different.
g and j are different.
j and j are identical.
<|im_end|>

<|im_start|>user
This is incorrect.
The answers should be:
s and none are different.
(There was no letter 2 steps ago.)
j and none are different.
(There was no letter 2 steps ago.)
g and s are different.
(The letter 2 steps ago was s.)
j and j are identical.
(The letter 2 steps ago was j.)
Now, given the sequence s, j, s, g,
what should the answers be?
<|im_end|>

<|im_start|>assistant
s and none are different.
j and s are different.
s and j are different.
g and s are different.
(The letter 2 steps ago was s.)
<|im_end|>
\end{verbatim}
}

\begin{figure*}[t]
  \includegraphics[width=1\linewidth]{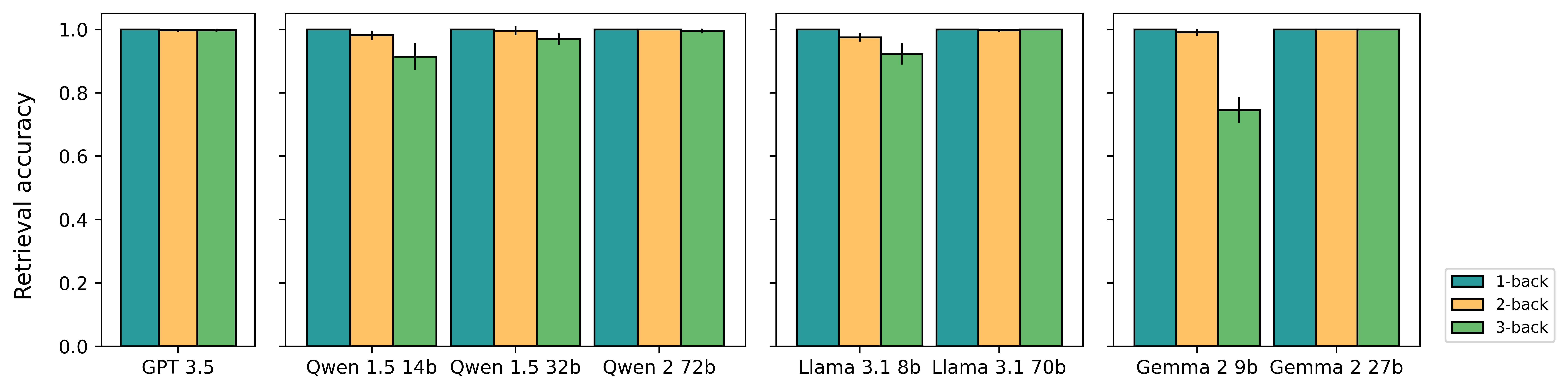}
  \caption {Average retrieval accuracies on 1-, 2-, and 3-back tasks, grouped by model family.}
  \label{fig:perf-v2}
\end{figure*}

\vspace*{0pt}
\section{Reciting \textit{N} Most Recent Stimuli}
\label{appendix:alt-prompt}

We experiment with an alternative answer format that encodes task requirements in greater detail. For 2-back trials, models are instructed to answer ``\textit{current: }\{\textit{current letter}\}\textit{, 1 back: }\{\textit{letter 1 back}\}\textit{, 2 back: }\{\textit{letter 2 back}\}\textit{; current letter }\{\textit{current letter}\}\textit{ and letter 2 back }\{\textit{letter 2 back}\}\textit{ are }\{\textit{different / identical}\}.'' The 3-back format is analogous.

Retrieval accuracies show significant improvements, including for T3 models, though their performances still lag slightly on 3-back trials (Figure~\ref{fig:perf-v2}). We include these results only for comparison, given that this format changes the original task into one that allows covert verbal rehearsal. In human experiments, participants would not have enough time to recite all $n$ most recent letters upon presentation of each new letter. However, these results do highlight the malleability of language models' performance on working memory tasks.

\end{document}